\documentclass{article} %
\usepackage{template}

\usepackage[utf8]{inputenc} %
\usepackage[T1]{fontenc}    %
\usepackage[backref=section]{hyperref}
\usepackage{url}

\usepackage{booktabs}
\usepackage{multirow}
\usepackage{colortbl}
\usepackage{multicol}
\usepackage{amsfonts}
\usepackage{nicefrac}
\usepackage{microtype}
\usepackage[most]{tcolorbox}
\usepackage[dvipsnames]{xcolor}
\usepackage{latexsym}
\usepackage{graphicx}
\usepackage{float}
\usepackage{subcaption}
\usepackage{wrapfig}
\usepackage{dirtytalk}
\usepackage{enumitem}
\usepackage{makecell}
\usepackage{titletoc}
\usepackage{minted}
\usepackage{todonotes}

\usepackage[toc,page,header]{appendix}

\usepackage{bm}

\usepackage{tabularx} 
\usepackage{arydshln}
\usepackage{ragged2e} 
\newcolumntype{L}{>{\RaggedRight\hangafter=1\hangindent=0em}X}

\usepackage{amsmath}
\usepackage{amssymb}
\usepackage{mathtools}
\usepackage{amsthm}

\usepackage[linesnumbered,ruled,vlined]{algorithm2e}

\hypersetup{
    colorlinks=true,
    linkcolor=Maroon,
    citecolor=ForestGreen,
    filecolor=magenta,      
    urlcolor=magenta,
}

\usepackage[capitalize,noabbrev]{cleveref}
\crefname{section}{§}{§§}
\Crefname{section}{§}{§§}
\crefname{lemma}{lemma}{lemma}
\Crefname{lemma}{Lemma}{Lemma}

\usepackage{calligra}
\DeclareMathAlphabet{\mathcalligra}{T1}{calligra}{m}{n}

\usepackage{pifont}

\theoremstyle{plain}

\theoremstyle{definition}

\theoremstyle{remark}

\renewcommand{\paragraph}[1]{\vspace{1mm}\noindent\textbf{#1}}

\DeclareCaptionLabelFormat{cont}{#1~#2\alph{ContinuedFloat}}
\tcbset{
  promptbox/.style={
    top=10pt,
    colback=lightgray!20,
    colframe=Black,
    colbacktitle=NavyBlue,
    enhanced,
    center,
    attach boxed title to top center={yshift=-0.1in,xshift=0.0in},
    boxed title style={boxrule=0pt,colframe=white,},
  }
}
\newtcolorbox{promptbox}[2][]{promptbox, title=#2,#1}
\tcbset{
  takeawaybox/.style={
    top=10pt,
    colback=lightgray!20,
    colframe=Black,
    colbacktitle=BurntOrange,
    enhanced,
    center,
    attach boxed title to top center={yshift=-0.1in,xshift=0.0in},
    boxed title style={boxrule=0pt,colframe=white,},
  }
}
\newtcolorbox{takeawaybox}[2][]{takeawaybox, title=#2,#1}
\tcbset{
  observationbox/.style={
    top=10pt,
    colback=lightgray!20,
    colframe=Black,
    colbacktitle=YellowGreen,
    enhanced,
    center,
    attach boxed title to top center={yshift=-0.1in,xshift=0.0in},
    boxed title style={boxrule=0pt,colframe=white,},
  }
}
\newtcolorbox{observationbox}[2][]{observationbox, title=#2,#1}

\newtcbtheorem[number within=section]{prompt}{Prompt}{
    top=10pt,
    colback=lightgray!10,
    colframe=Black,
    colbacktitle=NavyBlue,
    coltitle=white,
    enhanced,
    center,
    fonttitle=\bfseries,
    breakable,
    separator sign={:},
    description delimiters={}{},
    boxrule=0.8pt,
    arc=3pt,
    outer arc=3pt,
    boxsep=0pt,
    left=12pt,right=12pt,top=10pt,bottom=10pt,
    lefttitle=12pt,righttitle=12pt,toptitle=5pt,bottomtitle=5pt,
    before skip=12pt,after skip=12pt,
    pad at break*=3mm,
    fontupper=\normalsize
}{pt}

\definecolor{CaseFrame}{HTML}{4C4C4C}
\definecolor{CaseSummary}{HTML}{F0F0F0}
\definecolor{CaseGreen}{HTML}{188B35}
\definecolor{CaseGreenLight}{HTML}{F0FCF1}
\definecolor{CaseExecutorA}{HTML}{DDF5DF}
\definecolor{CaseExecutorB}{HTML}{FFF0E4}

\newtcbtheorem[number within=section]{case}{Case}{
    enhanced,
    breakable,
    colback=white,
    colframe=CaseFrame,
    colbacktitle=CaseFrame,
    coltitle=white,
    fonttitle=\bfseries,
    separator sign={:},
    description delimiters={}{},
    boxrule=0.8pt,
    arc=3pt,
    outer arc=3pt,
    boxsep=0pt,
    left=12pt,right=12pt,top=10pt,bottom=10pt,
    lefttitle=12pt,righttitle=12pt,toptitle=5pt,bottomtitle=5pt,
    before skip=12pt,after skip=12pt,
    pad at break*=3mm,
    fontupper=\normalsize
}{cs}

\newtcolorbox{casesummary}{
    enhanced,frame hidden,colback=CaseSummary,
    sharp corners,boxrule=0pt,boxsep=0pt,
    left=9pt,right=9pt,top=7pt,bottom=7pt,
    before skip=0pt,after skip=7pt
}

\newtcolorbox{caseoutcome}{
    enhanced,frame hidden,colback=white,
    borderline west={1.5pt}{0pt}{CaseFrame!35},
    sharp corners,boxrule=0pt,boxsep=0pt,
    left=7pt,right=3pt,top=2pt,bottom=2pt,
    fontupper=\small,
    before skip=4pt,after skip=3pt
}

\newtcolorbox{caseanalysis}{
    enhanced,
    title=Analysis,
    colframe=CaseGreen,colback=CaseGreenLight,
    colbacktitle=CaseGreenLight,coltitle=CaseGreen,
    fonttitle=\bfseries,fontupper=\small,
    boxrule=0.6pt,titlerule=0.4pt,
    arc=3pt,boxsep=0pt,
    left=9pt,right=9pt,top=6pt,bottom=6pt,
    lefttitle=9pt,toptitle=3pt,bottomtitle=3pt,
    before skip=9pt,after skip=0pt
}

\newcommand{\casemilestone}[1]{%
    \par\addvspace{7pt}\noindent{\small\bfseries #1}\par\nobreak\vspace{3pt}}
\newcommand{\caseexecutora}{\cellcolor{CaseExecutorA}\texttt{model\_a}}
\newcommand{\caseexecutorb}{\cellcolor{CaseExecutorB}\texttt{model\_b}}
\providecommand{\casepageseparator}{\par}

\newtcbtheorem[number within=section]{tool}{Tool Card}{
    top=10pt,
    colback=lightgray!20,
    colframe=Black,
    colbacktitle=YellowGreen,
    coltitle=white,
    enhanced,
    center,
    fonttitle=\bfseries,
    breakable,
    separator sign={:},
    description delimiters={}{},
    boxrule=0.8pt,
    arc=3pt,
    outer arc=3pt,
    boxsep=0pt,
    left=12pt,right=12pt,top=10pt,bottom=10pt,
    lefttitle=12pt,righttitle=12pt,toptitle=5pt,bottomtitle=5pt,
    before skip=12pt,after skip=12pt,
    pad at break*=3mm,
    fontupper=\normalsize
}{tl} 

\lstdefinelanguage{yaml}{
  keywords={true,false,null,y,n},
  keywordstyle=\color{blue}\bfseries,
  basicstyle=\ttfamily\small,
  commentstyle=\color{gray}\itshape,
  stringstyle=\color{purple},
  moredelim=[l][\color{orange}]{\&},
  moredelim=[l][\color{magenta}]{\*},
  comment=[l]{\#},
  morecomment=[s]{/*}{*/},
  morestring=[b]',
  morestring=[b]"
} 

\newcommand{\PromptLine}[1]{\noindent\detokenize{#1}\par}
\newcommand{\PromptIndent}[1]{\noindent\hspace*{1.5em}\detokenize{#1}\par}
\newcommand{\PromptBlank}{\vspace{0.55em}}
\newcommand{\Boxed}[1]{\textbackslash boxed\{\detokenize{#1}\}}
\newcommand{\PromptRaw}[1]{\noindent #1\par}

\usepackage{xspace}
\newcommand{\name}{%
  \texorpdfstring{\textsc{E$^{\text{3}}$-Orch}}{E3-Orch}\xspace%
}

\definecolor{carnelian}{rgb}{0.7, 0.11, 0.11}

\title{\name: Towards Effective, Efficient, and Extensible Agentic Orchestration with Reinforcement Learning}

\author{Junnan Liu, Linhao Luo, Thuy-Trang Vu \& Gholamreza Haffari  \\
Department of Data Science and AI, Faculty of Information Technology, \\
Monash University, Australia\\
}

\date{\today}
\correspondence{Junnan Liu at
  \href{mailto:junnan.liu@monash.edu}{\texttt{junnan.liu@monash.edu}}}

\begin{document}

\maketitle
\begin{abstract}
Agentic orchestration enables multiple autonomous agents to collaborate on complex tasks through adaptive task decomposition, delegation and execution.
However, existing orchestrators largely rely on hand-crafted orchestration logic and prompting strategies, limiting their ability to adapt and generalise across tasks and executor configurations.
In this work, we propose \name, a reinforcement learning framework for learning effective, efficient, and extensible agentic orchestration through a \emph{milestone-plan-act} workflow.
Rather than planning all subtasks upfront, \name organises execution around \emph{milestones}, a mid-level abstraction that scopes planning and execution around meaningful intermediate objectives, enabling the orchestrator to adapt its decisions as the task evolves.
For each milestone, the orchestrator constructs a dependency-aware plan, delegates subtasks to suitable executors, and parallelises independent subtasks for efficient execution.
We learn the orchestration policy from execution feedback, using milestone and plan decisions as natural units for fine-grained credit assignment.
To estimate their downstream contributions, we construct tree-structured rollouts that compare alternative decisions under shared execution histories.
The policy is jointly optimised for task performance and efficiency through complementary rewards for performance, execution cost, and planning completeness, including an uncertainty-aware performance reward that accounts for stochastic downstream outcomes.
Across seven benchmarks spanning diverse domains, \name achieves the best task performance under different agent configurations, improving on the strongest baselines by $0.7$-$3.8$ points, and achieves $1.16$-$1.59\times$ higher intelligence efficiency. 
The learned policy also transfers to new agent configurations introduced only at evaluation time, supporting the extensibility of agentic orchestration.

\end{abstract}

\section{Introduction}
\label{sec:introduction} 
Recent advances in large language model (LLM) agents have enabled multi-agent systems (MAS) that combine complementary capabilities to tackle complex tasks through reasoning, tool use, and environment interaction~\citep{YaoZYDSN023,ShinnCGNY23,QinLYZYLLCTQZHT24,ErdoganL0MFAKG25}.
Many early MAS rely on task allocations and execution workflows specified before execution, limiting their ability to adapt as tasks evolve during execution~\citep{abs-2308-08155,HongZCZCWZWYLZR24,QianLLCDL0CSCXL24,Du00TM24,abs-2505-14996,abs-2505-19591,abs-2603-22386}.
Recent work has therefore shifted toward \emph{agentic orchestration}, moving from fixed workflows to adaptive, execution-aware coordination of agents and tools. This paradigm has been adopted in advanced agentic systems, including Claude Code agent teams~\citep{anthropic2026agentteams}, OpenAI's GPT-5.6 Sol Ultra~\citep{openai2026cdcprompt,openai2026cdcproof}, and Kimi K2.5 Agent Swarm~\citep{abs-2602-02276}.
However, existing orchestration policies still largely rely on hand-crafted logic and prompting strategies~\citep{LiuZLY24Dylan,ZhangYSWYF00C25,SuLXSTSH26,abs-2603-11445}. Their effectiveness therefore remains dependent on human-designed rules and the orchestration capabilities of general-purpose LLMs, potentially limiting adaptation to evolving task states and generalization across tasks and agent configurations.

\begin{wrapfigure}{r}{0.6\textwidth}
    \centering
    \includegraphics[width=0.58\textwidth]{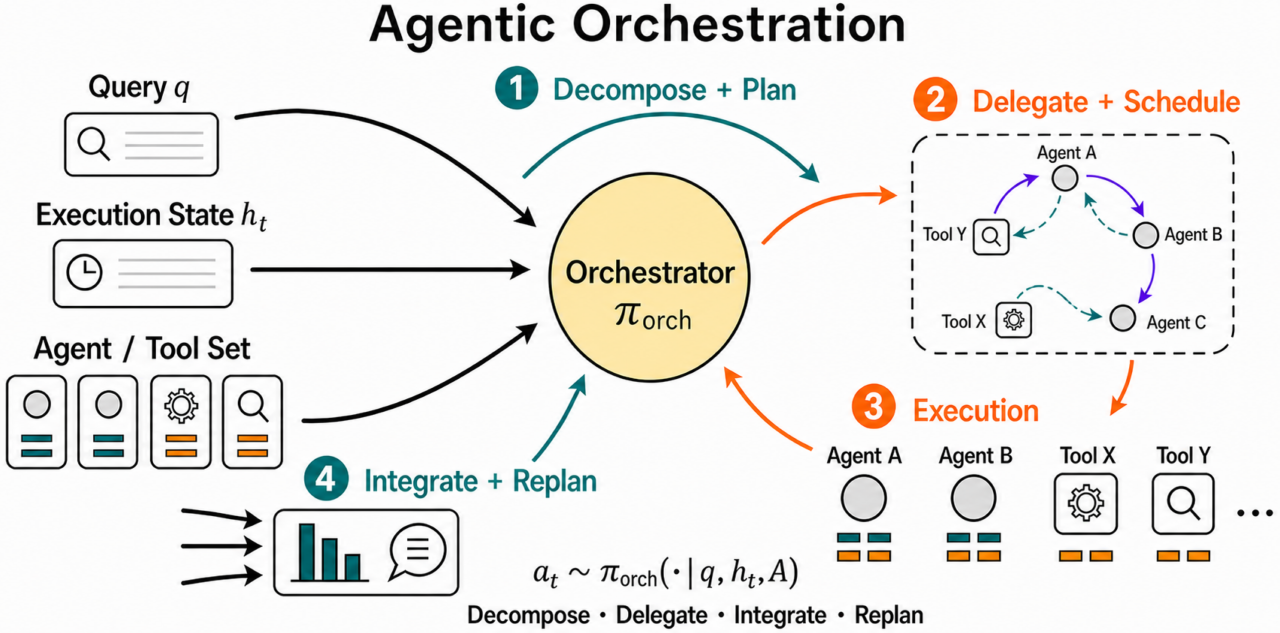}
    \caption{Agentic orchestration adaptively decomposes a task, delegates work, executes agents and tools, and replans from intermediate results.}
    \label{fig:intro}
    \vspace{-1.0em}
\end{wrapfigure}

As illustrated in \Cref{fig:intro}, an agentic orchestrator must dynamically decompose tasks, delegate subtasks, manage execution dependencies, and integrate intermediate outcomes to guide execution~\citep{abs-2603-22386,abs-2608-23283}.
To move beyond hand-crafted strategies, recent work has 
used reinforcement learning to learn orchestration policies from execution feedback to optimize for objectives such as agent selection, workflow construction, and execution cost~\citep{abs-2505-19591,YueZLWWCQ25,abs-2601-14652}. 
However, many existing approaches often operate at the level of atomic subtasks~\citep{abs-2505-14996,abs-2505-19591,YueZLWWCQ25}, either planning a complete workflow upfront~\citep{abs-2601-14652,SuLXSTSH26} or executing subtasks sequentially as the task progresses~\citep{abs-2510-05592}. 
This fine-grained formulation poses challenges for both orchestration and learning. Planning a complete workflow upfront requires long-horizon reasoning and limits adaptation to intermediate outcomes, while sequential execution can obscure dependency structure and forgo opportunities for parallel execution~\citep{KimMTLMKG24,abs-2603-22386}.
At the same time, it further complicates credit assignment, as long trajectories of fine-grained decisions make it difficult to attribute final outcomes to individual decisions~\citep{abs-2510-05592}, particularly under stochastic downstream execution.
Moreover, orchestration must balance multiple objectives, including task performance, execution efficiency, and planning completeness. These challenges motivate a higher-level orchestration abstraction that organizes planning and execution into manageable units, supports adaptive and dependency-aware execution, and provides meaningful units for credit assignment.

Effective orchestration requires the right abstraction for both execution and learning. We therefore organize execution around \emph{milestones}: meaningful intermediate objectives that structure a complex task into manageable stages, avoiding the need to plan a long sequence of atomic subtasks at once. 
Building on this abstraction, we propose \name (\underline{E}ffective, \underline{E}fficient, and \underline{E}xtensible Agentic \underline{Orch}estration), a reinforcement learning framework based on an iterative \emph{milestone-plan-act} workflow. 
For each milestone, the orchestrator constructs a dependency-aware plan that decomposes the milestone into subtasks and their dependencies~\citep{KimMTLMKG24,abs-2505-19591}, assigns them to suitable executors, and executes independent subtasks in parallel. 
The resulting outcomes condition the next milestone and plan, enabling the orchestration structure to be revised as execution proceeds. 
In this way, milestones scope planning and execution locally while providing natural points for adaptation, avoiding the need to commit to a complete workflow upfront.

Importantly, milestone and plan decisions also provide meaningful units for credit assignment. 
To estimate the contribution of each decision, we construct \emph{tree-structured rollouts}~\citep{hou2025treerl} that branch from shared execution histories, allowing alternative milestone and plan decisions to be compared through their downstream outcomes and execution costs. 
Based on these comparisons, we define three complementary decision-level rewards: a \emph{performance reward} that accounts for uncertainty in downstream outcome, a \emph{cost reward} that compares execution costs across alternative decisions, and a \emph{planning completeness reward} that assesses whether plans are well-formed and executable. 
We normalize these rewards to jointly optimize the orchestration policy for performance effectiveness, intelligence efficiency, and planning completeness.

We evaluate \name on seven benchmarks spanning mathematical reasoning, embodied decision-making, and deep research under two executor configurations.
Across both configurations, \name achieves the best task performance on each benchmark and yields $1.16$--$1.59\times$ higher intelligence efficiency than the strongest baseline. 
Further analyses provide evidence supporting the extensibility of profile-based orchestration.
Finally, ablations support the contributions of key components such as tree-structured rollouts to task performance and intelligence efficiency.  

In summary, our contributions are threefold: \textbf{(1) Orchestration abstraction.} We introduce a \emph{milestone-plan-act} workflow for adaptive, dependency-aware execution and meaningful credit assignment; \textbf{(2) Orchestration learning.} We develop tree-structured rollouts and multi-objective rewards for decision-level policy optimization under uncertain downstream outcomes; and \textbf{(3) Empirical performance.} We demonstrate consistent gains of \name in effectiveness, efficiency, and extensibility across diverse tasks and executor configurations.

\section{Related Work}\label{sec:related_work}

\vspace{-0.5em}

\paragraph{Language Agents and Multi-Agent Collaboration.}
Language agents augment LLM reasoning with planning, tool use, environmental feedback, and memory, which lets models solve interactive and long-horizon tasks~\citep{YaoZYDSN023,ShinnCGNY23,SchickDDRLHZCS23,QinLYZYLLCTQZHT24,ErdoganL0MFAKG25}.
Multi-agent systems distribute reasoning and execution across complementary workers through conversation, role specialization, and collaborative aggregation~\citep{abs-2308-08155,HongZCZCWZWYLZR24,QianLLCDL0CSCXL24,Du00TM24,QianXW0ZXDDC00025,WangWAZZ25}.
These systems broaden the available expertise and computation, but many of them fix roles, communication protocols, or execution pipelines before execution begins, and such structures are difficult to adapt once intermediate results change the remaining subtasks, their dependencies, or the workers best suited to solve them.

\vspace{-0.5em}

\paragraph{Agentic Orchestration.}
A second line of work therefore automates how agents and tools are composed, connected, and scheduled, spanning execution-graph and workflow optimization, task-specific agent composition and model routing, and runtime planning and replanning~\citep{ZhugeWKFKS24,ZhangYSWYF00C25,LiuZLY24Dylan,YueZLWWCQ25,abs-2505-14996,SuLXSTSH26,abs-2603-11445,abs-2603-22386}.
In parallel, supervision and reinforcement learning have been used to improve collaboration and orchestration policies rather than encoding them entirely in prompts or controller logic~\citep{sun2026agentpo,abs-2505-19591,abs-2601-14652,abs-2510-05592}.
Learning orchestration over long horizons nevertheless remains difficult: the execution structure must adapt to evolving task states, outcome-level feedback must be attributed at a meaningful decision granularity, and the three objectives must be optimized without obscuring their distinct signals.
We address these requirements by training an LLM orchestrator to construct and revise milestone-level execution graphs, using milestones as the unit of credit assignment and decoupled rewards for multi-objective optimization.

\section{Method} 
\label{sec:method}

\subsection{Preliminaries}\label{sec:preliminary} 

\vspace{-0.5em}

\paragraph{Agentic Orchestration.}
In this paper, we consider an agentic orchestration system that comprises an \emph{orchestrator} $\pi_{\text{orch}}$ with parameters $\theta$ and a pool of executors $\mathcal{A}=\{a_1,\ldots,a_K\}$. with heterogeneous operational profiles
Each executor $a_k$ is an LLM agent described by an operational profile $\phi_k=(\mathbf{c}_k,\kappa_k,\mathcal{T}_k)$, summarizing its capabilities, execution cost, and available tools.
The resulting profile pool $\Phi=\{\phi_1,\ldots,\phi_K\}$ can incorporate additional executor metadata without changing the orchestration workflow. 
Given a query $q$, the orchestrator proposes an intermediate objective as a milestone, decomposes it into subtasks with execution dependencies as a plan, and delegates each subtask to an executor chosen from $\Phi$.
A subtask runs once its dependencies are satisfied, so independent subtasks execute in parallel. 
The orchestrator then integrates their results and decides whether to initiate another milestone-plan-act round or terminate with a final answer. 
Repeating this loop produces a trajectory $\tau$ of orchestration decisions and execution outcomes. 

\vspace{-0.5em}

\paragraph{Training Objective.}
We update only the orchestrator parameters $\theta$, while the executors remain frozen and are treated as part of the execution environment.
An orchestrator is judged by the task performance $U(\tau)$ its trajectory achieves, the execution cost $C(\tau)$ it incurs, and the completeness $Q(\tau)$ of the decompositions and plans it produces, that is, whether they are well formed and executable.
Our objective is to maximize $U(\tau)$ and $Q(\tau)$ while minimizing $C(\tau)$.

\subsection{Milestone-Plan-Act Orchestration}\label{sec:method_design}

\begin{figure}[t]
    \centering
    \includegraphics[width=1.0\linewidth]{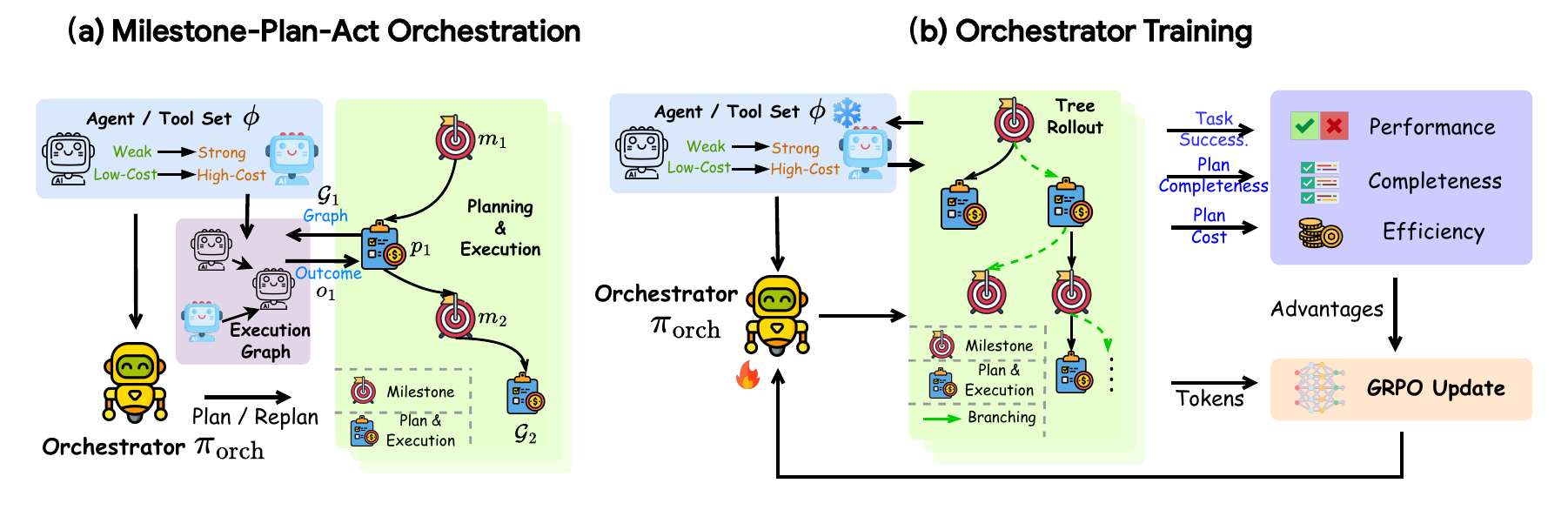}
    \caption{Overview of \name. \textbf{(a) Milestone-Plan-Act orchestration.} The orchestrator iteratively proposes milestones, plans dependency-aware execution graphs, delegates to agents and tools, and integrates outcomes for replanning. \textbf{(b) Orchestration training.} \name branches over orchestration  decisions to form tree-structured rollouts, evaluates alternative orchestration decisions using performance, cost, and completeness rewards.} 
    \label{fig:overview}
    \vspace{-1.5em}
\end{figure}

\vspace{-0.5em}
Task decomposition is central to orchestration, as it defines the units of work to be planned, scheduled, and executed.
However, planning all atomic subtasks in advance is difficult because intermediate outcomes can change their boundaries and dependencies, while long execution chains complicate agent scheduling and result integration.
\name therefore adopts an iterative \emph{milestone-plan-act} workflow centered on \emph{milestones}, i.e., intermediate objectives prompted to be concrete and independently verifiable, which scope task decomposition and parallel scheduling.
As illustrated in \Cref{fig:overview}~(a), each iteration $i$ proposes a milestone, plans and executes its subtasks, and aggregates their outputs to guide subsequent planning.

\paragraph{Milestone Proposal.}
Conditioned on the query $q$, the current execution history $h_{i-1}=(m_1,p_1,o_1,\ldots,m_{i-1},p_{i-1},o_{i-1})$, and the pool profile $\Phi$, the orchestrator proposes the next milestone $m_i\sim\pi_{\text{orch}}(\cdot\mid q,h_{i-1},\Phi)$ using the prompt in \Cref{app:milestone_prompt}.
Milestones are generated incrementally, allowing subsequent objectives to adapt to prior execution outcomes.

\vspace{-0.5em}
\paragraph{Plan Generation.}
The orchestrator then plans the selected milestone as $p_i\sim\pi_{\text{orch}}(\cdot\mid q,h_{i-1},m_i,\Phi)$ (\Cref{app:plan_prompt}), parsed into a directed acyclic graph $\mathcal{G}_i=(\mathcal{V}_i,\mathcal{E}_i)$~\citep{KimMTLMKG24,abs-2505-19591}.
Each node $v_{ij}\in\mathcal{V}_i$ pairs a subtask with its assigned executor $a(v_{ij})\in\mathcal{A}$, which may invoke tools from $\mathcal{T}_{a(v_{ij})}$ and return an output $o_{ij}$.
Each edge $(v_{ij},v_{ik})\in\mathcal{E}_i$ records that $v_{ik}$ consumes the output of $v_{ij}$.
Each plan thus jointly specifies the subtask decomposition, dependency structure, and executor assignments, enabling parallel execution subject to the milestone's dependency constraints.

\vspace{-0.5em}
\paragraph{Aggregation.}
The execution engine invokes every node whose dependencies are satisfied, running independent nodes in parallel, and aggregates the node outputs into a milestone-level outcome $o_i=\operatorname{Aggregate}(\{o_{ij}:v_{ij}\in\mathcal{V}_i\})$ using Prompt~\ref{pt:milestone_outcome_prompt}.
The orchestrator appends $o_i$ to the execution history and decides whether to initiate another orchestration round or terminate and answer $q$.
The complete execution trajectory is formulated as $\tau=(m_1,p_1,o_1,\ldots,m_n,p_n,o_n)$, while the milestone proposals $m_i$ and plan generations $p_i$ are the major explicit orchestration actions to be optimized while other orchestration decisions are implicitly learned during training.

\subsection{Orchestration Training}\label{sec:method_learning}

\subsubsection{Tree-Structured Orchestration Rollouts}\label{sec:tree_rollout}

\begin{wrapfigure}{l}{.7\textwidth}
    \centering
    \includegraphics[width=.68\textwidth]{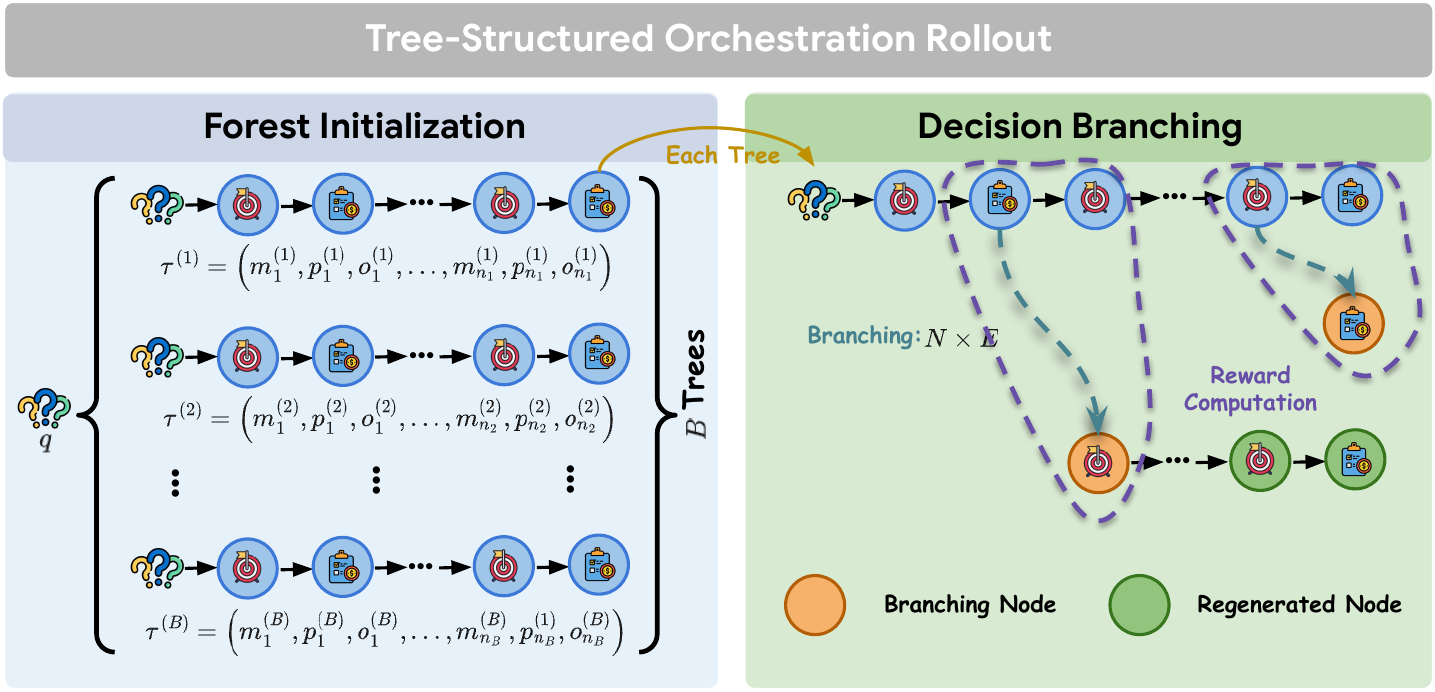}
    \vspace{-0.5em}
    \caption{Illustration of the tree-structured orchestration rollout used to train \name. The procedure produces up to $B\times N\times E$ newly expanded trajectories.}
    \label{fig:rollout}
    \vspace{-1.0em}
\end{wrapfigure}

A common RL training setup~\citep{abs-2402-03300} independently samples complete trajectories and uses terminal rewards to guide policy updates.
For orchestration decisions, differences in their terminal rewards may arise from different preceding histories, subsequent decisions, or stochastic executor behavior, making it difficult to isolate the decision contribution. 
We therefore adopt tree-structured rollouts~\citep{hou2025treerl} that branch at selected decisions under a shared execution history and regenerate downstream trajectories for each alternative decision, as illustrated in \Cref{fig:rollout}.
The resulting branches provide multiple downstream outcome samples for estimating the performance of intermediate decisions, while sibling branches sharing the same execution history provide matched alternatives for relative cost comparison.

\vspace{-0.5em}
\paragraph{Forest Initialization.}
For each query $q$ we sample $B$ complete trajectories in parallel, using the milestone-plan-act workflow described in \Cref{sec:method_design}. These sampled trajectories become the roots of $B$ trees as illustrated on the left of \Cref{fig:rollout}. 

\paragraph{Decision Branching. } 
Within each tree we randomly select $N$ orchestration decision nodes including both milestone and plan nodes as branching points. 
Branching keeps the history preceding the node, resamples the decision there, and regenerates every downstream decision and execution conditioned on the new outcomes. 
A milestone branch resamples the milestone together with its plan, whereas a plan branch holds the milestone fixed and regenerates only its execution graph. 
We choose to branch at milestones and plans to explore along the two orchestration decision dimensions of task decomposition and execution strategy, respectively. 
We repeat this expansion for $E$ rounds and add every valid branch to the forest, which yields up to $B\times N\times E$ new trajectories, as shown on right of \Cref{fig:rollout}.

\vspace{-0.5em}
\subsubsection{Rewards Design}\label{sec:reward_design}

\vspace{-0.5em}
Based on the tree-structured rollouts, we design three rewards to evaluate individual orchestration decisions in terms of task performance, intelligence efficiency, and planning completeness.
We use $\nu$ to denote a decision node representing either a milestone or a plan conditioned on the preceding execution history.

\vspace{-0.5em}
\paragraph{Performance Reward} evaluates an orchestration decision's task performance based on the outcomes of its downstream execution.
The downstream task performance is stochastic, as subsequent decisions and executor behavior may vary.
The empirical success rate alone does not capture the uncertainty arising from limited samples, particularly when decision nodes have different numbers of descendant trajectories.
We therefore design a probabilistic performance reward that accounts for uncertainty in each decision's success probability. 

Let $\hat{y}(\tau)$ be the final answer of trajectory $\tau$ and $y^*$ the reference answer.
We define the trajectory-level success indicator as $r_{\mathrm{perf}}(\tau)=\mathbb{I}[\operatorname{Eval}(\hat{y}(\tau);q,y^*)=1]$, which equals $1$ if the final answer is judged correct and $0$ otherwise. 
Here, $\operatorname{Eval}$ denotes the benchmark-specific correctness evaluator described in \Cref{sec:experimental_setup}. 
For a decision node $\nu$, let $n_\nu$ denote the number of descendant trajectories and $k_\nu$ the number of successful ones.
We place a uniform prior on its success probability $\psi(\nu)$ and update it using these observed outcomes.
To obtain a scalar reward, we compute the probability that this success probability exceeds that of a reference posterior corresponding to zero successes in the same number of attempts:  
\begin{equation} 
    \setlength\abovedisplayskip{1.3pt}%
    \setlength\belowdisplayskip{1.3pt}
    \begin{aligned} 
        & \psi_1\sim\operatorname{Beta}(k_{\nu}+1,n_{\nu}-k_{\nu}+1), \quad
        \psi_0\sim\operatorname{Beta}(1,n_{\nu}+1), \\
        & R_{\mathrm{perf}}(\nu) = 1- \frac{
            \mathrm{B}\!\left(k_\nu+1,\,2n_\nu-k_\nu+2\right)
            }{
            \mathrm{B}\!\left(k_\nu+1,\,n_\nu-k_\nu+1\right)
            },
    \end{aligned}
    \label{eq:performance_reward}
\end{equation} 
where $\mathrm{B}(\cdot,\cdot)$ denotes the beta function.
By comparing the posterior distributions, this reward incorporates uncertainty and quantifies the evidence that the decision outperforms a zero-success baseline \(\psi_0\).

\vspace{-0.5em}
\paragraph{Cost Reward} evaluates the downstream execution cost of an orchestration decision, corresponding to the cost component of intelligence efficiency. 
To control for differences in execution history, we compare costs among sibling decisions sharing the same execution state.  
For a decision node $\nu$, let $C(\nu)$ denote its downstream execution cost and $\mathcal{S}(\nu)$ denote its sibling set.
We define the cost reward by comparing this cost with the sibling average and normalizing the difference by the sibling standard deviation:
\begin{equation}
    \setlength\abovedisplayskip{1.3pt}%
    \setlength\belowdisplayskip{1.3pt}
    R_{\mathrm{cost}}(\nu)
    =
    \frac{
        \mu\!\left(\{C(\nu'): \nu'\in\mathcal{S}(\nu)\}\right)
        - C(\nu)
    }{
        \sigma\!\left(\{C(\nu'): \nu'\in\mathcal{S}(\nu)\}\right)
        + \delta
    },
    \label{eq:efficiency_reward}
\end{equation}
where $\mu(\cdot)$ and $\sigma(\cdot)$ denote the average and standard deviation of sibling costs, respectively, and $\delta>0$ is a small constant for numerical stability.
This reward is positive when a decision incurs lower downstream execution cost than the sibling average and negative vice versa. 

\vspace{-0.5em}
\paragraph{Completeness Reward} evaluates an orchestration decision's structural validity in terms of format validity and executability.
Format validity requires milestone and plan outputs to contain all required fields and conform to the expected schema. For plans, structural validity further requires all dependency references to be valid and acyclic, while assignment validity requires every step to name an executor in the available pool. A plan is executable when these constraints are satisfied and its execution successfully produces a milestone-level outcome.
We define the completeness reward as a binary indicator of whether both requirements are satisfied:
\begin{equation}
    \setlength\abovedisplayskip{0.0pt}%
    \setlength\belowdisplayskip{0.0pt}
    R_{\mathrm{comp}}(\nu)
    = \mathbb{I}\!\left[
        \nu\ \text{is well formed and executable}
    \right] \in\{0,1\}.
    \label{eq:completeness_reward}
\end{equation}
A decision receives a reward of $1$ when both requirements are met and $0$ otherwise, providing a validity signal at each decision node to discourage malformed or non-executable outputs during training.

\subsubsection{Policy Optimization}\label{sec:policy_optimization}

The three rewards are computed within its rollout tree. Each reward is then normalized across nodes of the same decision type in the entire forest, and the resulting advantages are combined to guide policy optimization.

Aggregating rewards before advantages normalization can obscure their individual learning signals.
We therefore compute advantages separately for each reward and combine them for GRPO optimization~\citep{abs-2601-05242}.
For $c\in\{\mathrm{perf},\mathrm{cost},\mathrm{comp}\}$, let $\mathcal{H}(\nu)$ denote the nodes of the same decision type as $\nu$ across all trees generated for query $q$. The advantage of node $\nu$ for reward $R_c$ is computed as:
\begin{equation}
    \setlength\abovedisplayskip{1.3pt}%
    \setlength\belowdisplayskip{1.3pt}
    A_c(\nu)
    =
    \frac{
        R_c(\nu) - \mu\!\left( \{R_c(\nu'): \nu'\in\mathcal{H}(\nu)\} \right)
    }{
        \sigma\!\left( \{R_c(\nu'): \nu'\in\mathcal{H}(\nu)\} \right) + \delta
    }.
    \label{eq:component_advantage}
\end{equation}
The final advantage for node $\nu$ is a weighted sum of the normalized advantages as 
\begin{equation}
\label{eq:combined_advantage}
A(\nu) = \lambda_{\mathrm{perf}}A_{\mathrm{perf}}(\nu) + \lambda_{\mathrm{cost}}A_{\mathrm{cost}}(\nu) + \lambda_{\mathrm{comp}}A_{\mathrm{comp}}(\nu),
\end{equation}
where $\lambda_{\mathrm{perf}}$, $\lambda_{\mathrm{cost}}$, and $\lambda_{\mathrm{comp}}$ are corresponding weights for each reward. 
Finally, the orchestrator is updated with the clipped GRPO objective~\citep{abs-2402-03300}. 
At decision node $\nu$, the state $s_\nu$ consists of the query, execution history, and executor profiles $\Phi$, while the action $z_\nu$ is the generated token sequence specifying a milestone or a plan.

\section{Experiments}

\begin{table}[t]
\centering
\caption{Task performance of \name and the baselines. The \textbf{Training} column indicates whether the orchestrator is learned. Results are averaged over 32 runs for AIME24/25 and 3 runs for the remaining benchmarks, where subscripts are standard deviations over those runs. Bold and underlined entries denote the best and second-best results within each executor block, respectively.}
\label{tab:math_results}
\resizebox{\linewidth}{!}{
\begin{tabular}{lccccccc}
\toprule
\multirow{2}{*}{\textbf{Method}} & \multirow{2}{*}{\textbf{Training}} & \multicolumn{2}{c}{\cellcolor{cyan!20} \textbf{Mathematical}} & \multicolumn{2}{c}{\cellcolor{Orchid!20} \textbf{Embodied}} & \multicolumn{2}{c}{\cellcolor{LimeGreen!20} \textbf{Deep Research}} \\
\cmidrule(lr){3-4} \cmidrule(lr){5-6} \cmidrule(lr){7-8}
& & \textbf{AIME24/25} & \textbf{MATH500} & \textbf{ALFWorld} & \textbf{WebShop} & \textbf{GAIA} & \textbf{HLE} \\
\midrule
\multicolumn{8}{c}{\cellcolor{gray!10} \includegraphics[height=.9em]{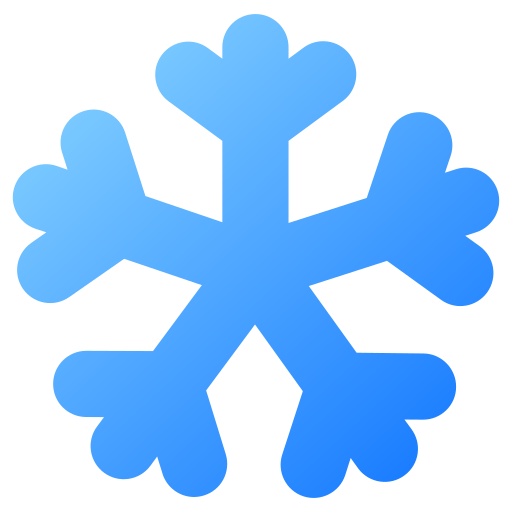} \textit{\textbf{Executor: Qwen2.5-7B-Instruct}}} \\
\midrule
\textsc{Direct Reasoning} & \textcolor{red!60!black}{\ding{56}} & $10.7_{\pm 0.3}$/$9.3_{\pm 0.4}$ & $72.1_{\pm 0.5}$ & $14.8_{\pm 0.4}$ & $8.1_{\pm 0.3}$ & $3.5_{\pm 0.2}$ & $1.1_{\pm 0.3}$ \\
\textsc{ReAct} & \textcolor{red!60!black}{\ding{56}} & $16.7_{\pm 0.4}$/$13.3_{\pm 0.5}$ & $77.0_{\pm 0.4}$ & $31.2_{\pm 0.6}$ & $19.5_{\pm 0.4}$ & $3.1_{\pm 0.3}$ & $1.5_{\pm 0.2}$ \\
\textsc{Reflexion} & \textcolor{red!60!black}{\ding{56}} & $22.3_{\pm 0.5}$/$18.7_{\pm 0.3}$ & $81.7_{\pm 0.6}$ & $42.7_{\pm 0.5}$ & $28.8_{\pm 0.5}$ & $4.5_{\pm 0.4}$ & $1.8_{\pm 0.3}$ \\
\hdashline
\textsc{AutoGen} & \textcolor{red!60!black}{\ding{56}} & $16.7_{\pm 0.3}$/$14.8_{\pm 0.4}$ & $79.5_{\pm 0.5}$ & $52.4_{\pm 0.4}$ & $41.9_{\pm 0.6}$ & $8.3_{\pm 0.5}$ & $2.7_{\pm 0.4}$ \\
\textsc{Multi-Agent Debate} & \textcolor{red!60!black}{\ding{56}} & $18.5_{\pm 0.4}$/$15.2_{\pm 0.3}$ & $80.1_{\pm 0.4}$ & $55.3_{\pm 0.7}$ & $44.7_{\pm 0.5}$ & $9.1_{\pm 0.4}$ & $2.9_{\pm 0.3}$ \\
\hdashline
\textsc{MAS-ZERO} & \textcolor{red!60!black}{\ding{56}} & $29.5_{\pm 0.6}$/$25.0_{\pm 0.5}$ & $85.1_{\pm 0.5}$ & $58.2_{\pm 0.6}$ & $49.7_{\pm 0.4}$ & $11.5_{\pm 0.5}$ & $3.0_{\pm 0.4}$ \\
\textsc{AgentFlow} & \textcolor{green!60!black}{\ding{52}} & $32.8_{\pm 0.5}$/\underline{$28.0_{\pm 0.4}$} & \underline{$88.1_{\pm 0.4}$} & $65.6_{\pm 0.5}$ & $57.3_{\pm 0.6}$ & $16.2_{\pm 0.4}$ & \underline{$6.6_{\pm 0.3}$} \\
\textsc{Puppeteer} & \textcolor{green!60!black}{\ding{52}} & $30.5_{\pm 0.4}$/$26.0_{\pm 0.5}$ & $85.9_{\pm 0.6}$ & $61.2_{\pm 0.4}$ & $54.1_{\pm 0.5}$ & $14.3_{\pm 0.6}$ & $4.1_{\pm 0.5}$ \\
\textsc{MAS-Orchestra} & \textcolor{green!60!black}{\ding{52}} & \underline{$33.5_{\pm 0.4}$}/$27.0_{\pm 0.6}$ & $86.7_{\pm 0.5}$ & \underline{$68.8_{\pm 0.5}$} & \underline{$58.9_{\pm 0.4}$} & \underline{$17.2_{\pm 0.5}$} & $3.5_{\pm 0.4}$ \\
\textsc{MasRouter} & \textcolor{green!60!black}{\ding{52}} & $28.2_{\pm 0.5}$/$25.3_{\pm 0.4}$ & $84.1_{\pm 0.4}$ & $63.1_{\pm 0.5}$ & $58.2_{\pm 0.5}$ & $15.1_{\pm 0.4}$ & $5.8_{\pm 0.5}$ \\
\textsc{DAAO} & \textcolor{green!60!black}{\ding{52}} & $30.2_{\pm 0.6}$/$26.0_{\pm 0.4}$ & $85.9_{\pm 0.5}$ & $60.1_{\pm 0.6}$ & $51.3_{\pm 0.4}$ & $12.1_{\pm 0.5}$ & $5.4_{\pm 0.4}$ \\
\hdashline
\name & \textcolor{green!60!black}{\ding{52}} & \textbf{35.0}$_{\pm \mathbf{0.4}}$/\textbf{31.5}$_{\pm \mathbf{0.5}}$ & \textbf{90.4}$_{\pm \mathbf{0.4}}$ & \textbf{70.2}$_{\pm \mathbf{0.5}}$ & \textbf{62.7}$_{\pm \mathbf{0.6}}$ & \textbf{20.5}$_{\pm \mathbf{0.4}}$ & \textbf{7.8}$_{\pm \mathbf{0.3}}$ \\
\midrule
\multicolumn{8}{c}{\cellcolor{gray!10} \includegraphics[height=.9em]{figs/snowflake.png} \textit{\textbf{Executor: GPT-5-mini}}} \\
\midrule
\textsc{Direct Reasoning} & \textcolor{red!60!black}{\ding{56}} & $48.3_{\pm 0.5}$/$40.1_{\pm 0.4}$ & $89.7_{\pm 0.3}$ & $84.3_{\pm 0.5}$ & $73.5_{\pm 0.6}$ & $26.5_{\pm 0.4}$ & $14.7_{\pm 0.5}$ \\
\textsc{ReAct} & \textcolor{red!60!black}{\ding{56}} & $54.2_{\pm 0.6}$/$45.1_{\pm 0.5}$ & $92.6_{\pm 0.4}$ & $87.1_{\pm 0.4}$ & $77.2_{\pm 0.5}$ & $29.1_{\pm 0.5}$ & $16.8_{\pm 0.4}$ \\
\textsc{Reflexion} & \textcolor{red!60!black}{\ding{56}} & $60.7_{\pm 0.4}$/$50.2_{\pm 0.6}$ & $94.8_{\pm 0.5}$ & $88.3_{\pm 0.6}$ & $79.1_{\pm 0.4}$ & $30.6_{\pm 0.3}$ & $17.9_{\pm 0.5}$ \\
\hdashline
\textsc{AutoGen} & \textcolor{red!60!black}{\ding{56}} & $56.0_{\pm 0.5}$/$47.3_{\pm 0.4}$ & $93.4_{\pm 0.4}$ & $89.5_{\pm 0.5}$ & $80.8_{\pm 0.5}$ & $31.7_{\pm 0.6}$ & $18.4_{\pm 0.4}$ \\
\textsc{Multi-Agent Debate} & \textcolor{red!60!black}{\ding{56}} & $58.1_{\pm 0.4}$/$49.0_{\pm 0.5}$ & $94.1_{\pm 0.5}$ & $90.2_{\pm 0.4}$ & $81.9_{\pm 0.6}$ & $32.5_{\pm 0.4}$ & $18.9_{\pm 0.5}$ \\
\hdashline
\textsc{MAS-ZERO} & \textcolor{red!60!black}{\ding{56}} & $61.5_{\pm 0.5}$/$52.0_{\pm 0.4}$ & $95.1_{\pm 0.4}$ & $90.2_{\pm 0.5}$ & $82.3_{\pm 0.5}$ & $33.5_{\pm 0.5}$ & $19.1_{\pm 0.4}$ \\
\textsc{AgentFlow} & \textcolor{green!60!black}{\ding{52}} & \underline{$69.4_{\pm 0.4}$}/$59.0_{\pm 0.5}$ & $97.1_{\pm 0.3}$ & \underline{$93.2_{\pm 0.4}$} & $86.5_{\pm 0.5}$ & $42.4_{\pm 0.5}$ & $23.6_{\pm 0.4}$ \\
\textsc{Puppeteer} & \textcolor{green!60!black}{\ding{52}} & $67.2_{\pm 0.5}$/$56.0_{\pm 0.6}$ & $96.3_{\pm 0.4}$ & $92.5_{\pm 0.6}$ & $85.1_{\pm 0.4}$ & $40.1_{\pm 0.4}$ & $22.4_{\pm 0.5}$ \\
\textsc{MAS-Orchestra} & \textcolor{green!60!black}{\ding{52}} & $68.5_{\pm 0.6}$/\underline{$61.5_{\pm 0.4}$} & \underline{$97.3_{\pm 0.3}$} & $92.0_{\pm 0.5}$ & \underline{$87.9_{\pm 0.4}$} & \underline{$43.1_{\pm 0.5}$} & \underline{$24.1_{\pm 0.3}$} \\
\textsc{MasRouter} & \textcolor{green!60!black}{\ding{52}} & $58.2_{\pm 0.4}$/$48.6_{\pm 0.5}$ & $94.1_{\pm 0.5}$ & $88.9_{\pm 0.4}$ & $79.9_{\pm 0.5}$ & $31.2_{\pm 0.5}$ & $18.1_{\pm 0.6}$ \\
\textsc{DAAO} & \textcolor{green!60!black}{\ding{52}} & $62.3_{\pm 0.5}$/$53.1_{\pm 0.4}$ & $95.5_{\pm 0.4}$ & $91.0_{\pm 0.5}$ & $83.1_{\pm 0.4}$ & $34.2_{\pm 0.6}$ & $19.5_{\pm 0.4}$ \\
\hdashline
\name & \textcolor{green!60!black}{\ding{52}} & \textbf{73.2}$_{\pm \mathbf{0.5}}$/\textbf{63.8}$_{\pm \mathbf{0.4}}$ & \textbf{98.0}$_{\pm \mathbf{0.2}}$ & \textbf{95.1}$_{\pm \mathbf{0.4}}$ & \textbf{89.4}$_{\pm \mathbf{0.4}}$ & \textbf{45.7}$_{\pm \mathbf{0.5}}$ & \textbf{26.2}$_{\pm \mathbf{0.4}}$ \\
\bottomrule
\end{tabular}
}
\vspace{-0.8em}
\end{table} 

\subsection{Experimental Setup} 
\label{sec:experimental_setup}

\vspace{-0.5em}
\paragraph{Datasets.}
We evaluate \name on seven benchmarks spanning three task families:
\textit{mathematical reasoning} (AIME24, AIME25, MATH500~\citep{HendrycksBKABTS21}),
\textit{embodied decision-making} (ALFWorld~\citep{ShridharYCBTH21}, WebShop~\citep{Yao0YN22}), and
\textit{deep research} (GAIA~\citep{MialonF0LS24}, Humanity's Last Exam (HLE)~\citep{abs-2501-14249}).
For training, we collect approximately 7,000 instances from DAPO-Math~\citep{abs-2503-14476}, OpenScienceReasoning-2~\citep{open_science_reasoning_2}, ALFWorld, WebShop, Natural Questions~\citep{KwiatkowskiPRCP19}, HotpotQA~\citep{Yang0ZBCSM18}, and WebDancer~\citep{abs-2505-22648} (\Cref{app:implementation_details}).

\vspace{-0.5em}
\paragraph{Baselines.}
We compare against three groups of systems:
\textit{single-agent methods} (\textsc{Direct Reasoning}, \textsc{ReAct}~\citep{YaoZYDSN023}, \textsc{Reflexion}~\citep{ShinnCGNY23});
\textit{multi-agent collaboration methods} (\textsc{AutoGen}~\citep{abs-2308-08155}, \textsc{Multi-Agent Debate}~\citep{Du00TM24}); and
\textit{agentic orchestration methods} (\textsc{MAS-ZERO}~\citep{abs-2505-14996}, \textsc{DAAO}~\citep{SuLXSTSH26}, \textsc{AgentFlow}~\citep{abs-2510-05592}, \textsc{Puppeteer}~\citep{abs-2505-19591}, \textsc{MAS-Orchestra}~\citep{abs-2601-14652}, \textsc{MasRouter}~\citep{YueZLWWCQ25}).
All methods share the same executor backbones and tool interfaces, and every baseline that includes an orchestrator uses the same orchestrator backbone and training settings as \name. 
Specifically, the rollout group size is aligned with that in the tree-structured rollout used by \name. 
See \Cref{app:baseline_details,app:baselines_implementation} for more details and differences with \name.

\begin{figure}[t]
\centering
\includegraphics[width=.95\linewidth]{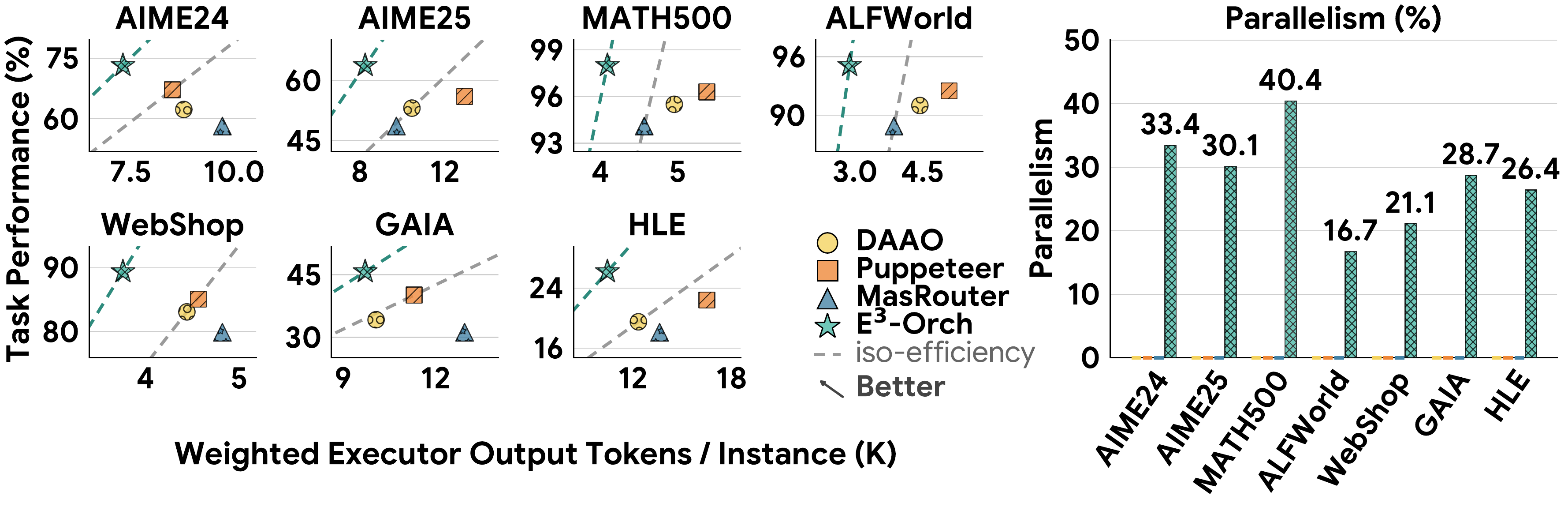}
\vspace{-0.5em}
\caption{Intelligence efficiency and execution parallelism.
\textbf{(Left)} \name achieves the Pareto frontier in intelligence efficiency with the highest task performance and lowest token cost.
\textbf{(Right)} \name executes $16.7$--$40.4\%$ of its calls in parallel, compared with $0\%$ for the evaluated baselines for superior execution efficiency.}
\label{fig:efficiency}
\vspace{-2.0em}
\end{figure}

\vspace{-0.5em}
\paragraph{Implementation Details.}
We instantiate the orchestrator $\pi_{\text{orch}}$ with Qwen2.5-7B-Instruct~\citep{abs-2412-15115} and train it with GRPO on A100 GPUs using the rLLM framework~\citep{rllm2025}. 
We summarize the training hyperparameters in \Cref{tab:train_param}.
The executor pool uses Qwen2.5-7B-Instruct during training, and at evaluation time we additionally consider a GPT-5-mini configuration~\citep{gpt5}.
Executors are exposed to the orchestrator through two anonymous profiles that differ in both capability and cost, follow \textsc{ReAct}~\citep{YaoZYDSN023}, and share a common tool library (\Cref{app:tools}).
We verify answers with Math-Verify~\citep{math_verify} for mathematical tasks and CompassVerifier~\citep{LiuLLXGLGZWZC25} otherwise (\Cref{app:verifier_prompt}).

\subsection{Main Results} \label{sec:main_results}

\vspace{-0.5em}
\paragraph{Performance Effectiveness.}
\name achieves the best result on each benchmark under both executor configurations (\Cref{tab:math_results}), improving on the strongest competing result in each column by $0.7$ to $3.8$ points.
The gains span mathematical reasoning, embodied decision-making, and deep research, which indicates that the learned orchestration policy is not tied to a single task structure.
By replacing Qwen2.5-7B-Instruct executors with GPT-5-mini, we observe that every method's absolute performance improves, while \name remains the best in every column. This indicates the learned orchestration ability can be transferred to a new executor configuration without any change to the milestone-plan-act interface. 
Scaling behavior across 3B and 7B orchestrator backbones is reported in \Cref{app:orchestrator_scaling}. 

\begin{wrapfigure}{r}{0.7\textwidth}
    \centering
    \vspace{-0.5em}
    \includegraphics[width=0.7\textwidth]{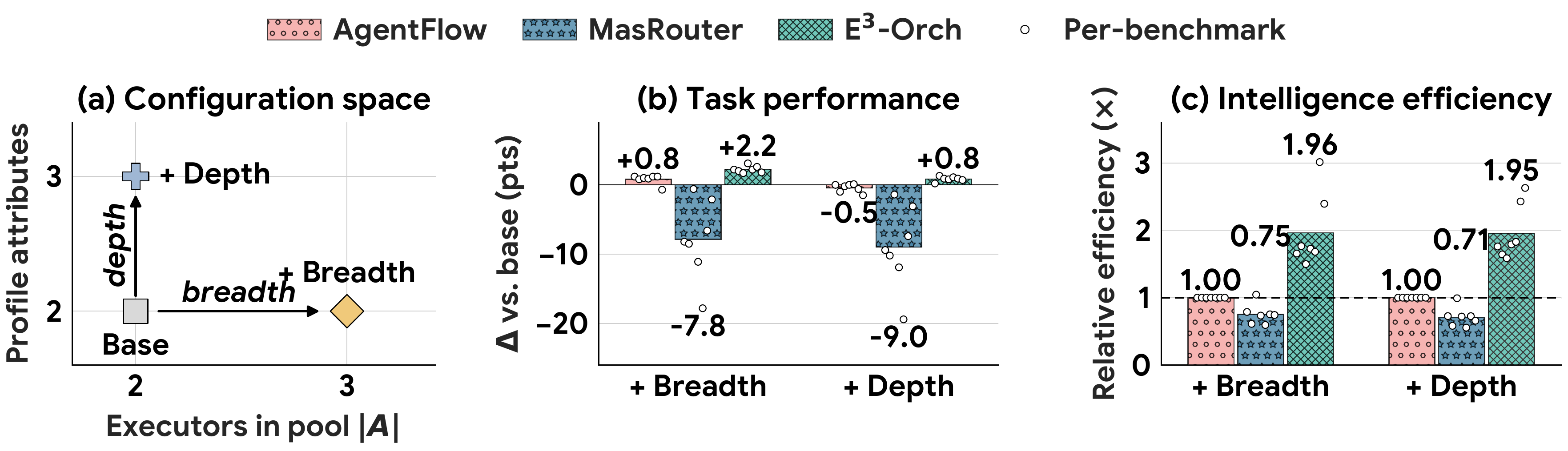}
    \caption{Extension across executor configurations. (\textbf{Left}) \emph{Breadth} adds a third executor (\Cref{tab:model_list_genel}) and \emph{depth} adds a capability attribute to each profile (\Cref{tab:model_desc_genel}). (\textbf{Middle}) Task performance change relative to the base configurations.(\textbf{Right}) intelligence efficiency relative to AgentFlow (dashed line). Bars show means across benchmarks, and dots show individual benchmark results. }
    \label{fig:executor_generalization}
    \vspace{-0.5em}
\end{wrapfigure}
\vspace{-0.5em} 

\vspace{-0.5em}
\paragraph{Intelligence Efficiency.}
We measure intelligence efficiency as task performance per cost-weighted executor output token, reflecting the goal of completing tasks accurately with fewer tokens~\citep{openai2026gpt56frontier} (see details in \Cref{app:intelligence_efficiency}).
As shown on the left of \Cref{fig:efficiency}, \name achieves the Pareto frontier with the highest task performance and lowest weighted token cost among the methods shown on all seven benchmarks. Its performance per token is $1.16$--$1.59\times$ higher than that of the best baseline, indicating the superior intelligence efficiency.
As shown on the right of \Cref{fig:efficiency}, \name executes $16.7$--$40.4\%$ of its executor calls in parallel, compared with $0\%$ for the evaluated baselines.
This suggests that the learned orchestrator can identify independently executable subtasks through task decomposition and organize them for concurrent execution.
Overlapping independent work can shorten task execution time, consistent with the lower wall-clock times reported in \Cref{app:wall_clock_time}. 

\paragraph{Executor Extensibility.}
We evaluate the extensibility of \name along two dimensions: \emph{breadth}, which adds a third executor to the pool, and \emph{depth}, which adds a capability attribute to each executor profile (\Cref{fig:executor_generalization}).
Under both extensions, \name improves task performance relative to its base configuration and achieves $\approx\!1.95\times$ intelligence efficiency compared to AgentFlow.
By contrast, MasRouter loses $7.8$--$9.0$ performance points relative to its base configuration and achieves only $0.71$--$0.75\times$ AgentFlow's efficiency.
These results suggest that \name can benefit from both a broader executor pool and more detailed capability profiles. Detailed extensibility results can be found in \Cref{app:general_executor}.

\subsection{Ablation Study}

\begin{table}[t]
\centering
\caption{Ablation results with Qwen2.5-7B-Instruct executors, averaged within each benchmark family. We report task performance ($\uparrow$) and token cost ($\downarrow$), the weighted executor output tokens per instance in thousands, their ratio is the intelligence efficiency of \Cref{fig:efficiency}.}
\label{tab:ablation}
\resizebox{\linewidth}{!}{
\begin{tabular}{lcccccc}
\toprule
\multirow{2}{*}{\textbf{Method}} & \multicolumn{2}{c}{\cellcolor{cyan!20} \textbf{Mathematical}} & \multicolumn{2}{c}{\cellcolor{Orchid!20} \textbf{Embodied}} & \multicolumn{2}{c}{\cellcolor{LimeGreen!20} \textbf{Deep Research}} \\
& \textbf{Perf.} & \textbf{Tokens (K)} & \textbf{Perf.} & \textbf{Tokens (K)} & \textbf{Perf.} & \textbf{Tokens (K)} \\
\midrule
\name & \textbf{52.3}$_{\pm \mathbf{0.4}}$ & \textbf{6.2}$_{\pm \mathbf{0.5}}$ & \textbf{66.5}$_{\pm \mathbf{0.5}}$ & \textbf{3.3}$_{\pm \mathbf{0.4}}$ & \textbf{14.2}$_{\pm \mathbf{0.4}}$ & \textbf{10.4}$_{\pm \mathbf{0.4}}$ \\
\midrule
\textit{w/} Standard Rollouts & $48.8_{\pm 0.5}$ & $7.8_{\pm 0.9}$ & $62.3_{\pm 0.6}$ & $4.4_{\pm 0.6}$ & $10.0_{\pm 0.3}$ & $11.8_{\pm 1.0}$ \\
\textit{w/o} Decision Branching & $49.2_{\pm 0.4}$ & $7.3_{\pm 0.6}$ & $65.1_{\pm 0.5}$ & $4.8_{\pm 0.4}$ & $12.5_{\pm 0.5}$ & $11.3_{\pm 0.5}$ \\
\textit{w/} Mean Performance Reward & $50.1_{\pm 0.6}$ & $6.5_{\pm 0.3}$ & $63.2_{\pm 0.4}$ & $3.3_{\pm 0.3}$ & $12.5_{\pm 0.4}$ & $10.9_{\pm 0.9}$ \\
\textit{w/o} Cost Reward & $50.6_{\pm 0.5}$ & $11.3_{\pm 0.9}$ & $65.1_{\pm 0.6}$ & $5.7_{\pm 0.6}$ & $13.5_{\pm 0.6}$ & $12.9_{\pm 0.6}$ \\
\textit{w/} Scalarized Reward & $50.5_{\pm 0.4}$ & $7.2_{\pm 0.6}$ & $64.5_{\pm 0.5}$ & $3.5_{\pm 0.4}$ & $12.9_{\pm 0.5}$ & $10.9_{\pm 0.9}$ \\
\bottomrule
\end{tabular}
}
\vspace{-1.0em}
\end{table} 

\Cref{tab:ablation} isolates the contributions of the rollout structure, the individual rewards, and the decoupled normalization, reporting task performance and token cost separately so that each component's effect on intelligence efficiency is visible. Additional analyses of the orchestration abstraction and completeness reward are provided in \Cref{app:analysis_orchestration,app:ablation_completeness}.
(1) \emph{Replacing tree-structured rollouts with independent standard rollouts} is the most damaging ablation on both axes, retaining only $0.62$--$0.74\times$ the intelligence efficiency of the full method, because without shared prefixes neither reward can be computed against a matched alternative.
(2) \emph{Removing decision branching} is milder ($0.80$--$0.85\times$), confirming that branching at milestone and plan decisions supplies useful variation beyond that of independent initial trajectories alone.
The two outcome-based rewards then affect complementary axes.
(3) \emph{Replacing the uncertainty-aware performance reward in \Cref{eq:performance_reward} with the mean performance of descendant trajectories} costs $1.7$--$3.3$ performance points while leaving token cost almost untouched, consistent with its motivation that nodes supported by different numbers of descendants should not receive identical confidence.
(4) \emph{Removing $R_{\mathrm{cost}}$} instead causes the smallest performance drop of any ablation yet inflates token cost by $24$--$82\%$, lowering intelligence efficiency to $0.53$--$0.77\times$: task performance alone would rank this variant among the best ablations, whereas intelligence efficiency ranks it last.
(5) Finally, \emph{combining the three raw rewards before normalizing once} keeps most of the cost benefit but pays $1.3$--$2.0$ performance points for it, falling below \emph{w/o Cost Reward} on task performance in every family, a pattern consistent with reward-signal collapse~\citep{abs-2601-05242}, where aggregation before normalization makes distinct reward combinations harder for the policy to distinguish.

\subsection{Behavior Analysis of the Learned Orchestrator}

\begin{figure}[t]
\centering
\includegraphics[width=.9\linewidth]{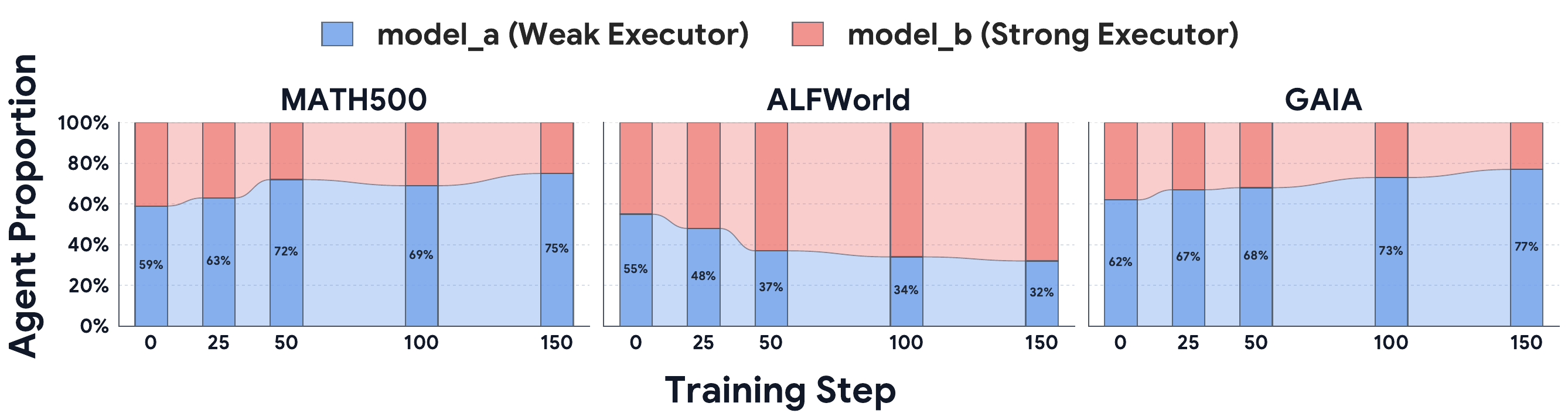}
\vspace{-0.5em}
\caption{Executor usage during training. \texttt{model\_a} is a weaker, lower-cost executor, whereas \texttt{model\_b} is a stronger and more expensive executor. }
\label{fig:agent_usage}
\vspace{-1.0em}
\end{figure} 

\begin{figure}[t]
\centering
\includegraphics[width=.95\linewidth]{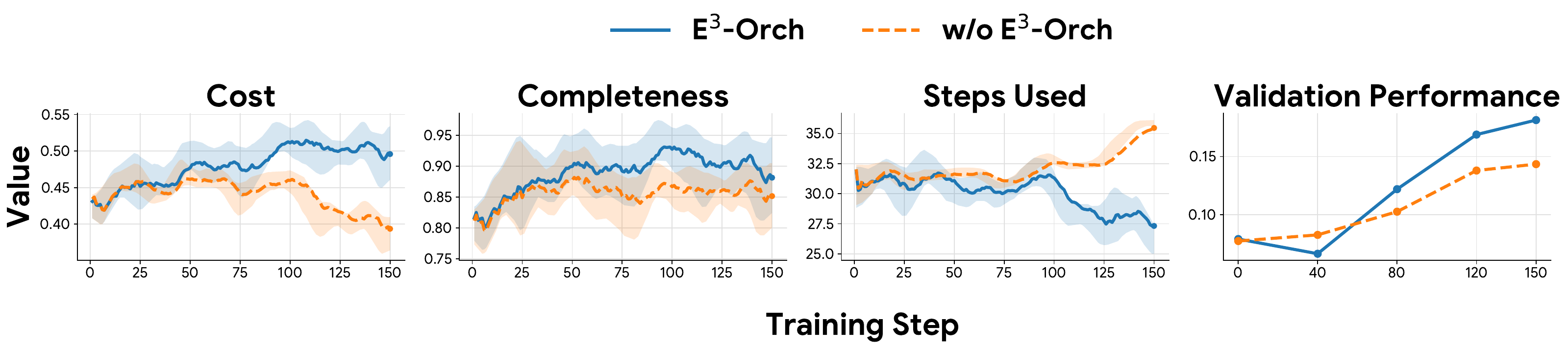}
\vspace{-1em}
\caption{Training dynamics of \name and the standard-GRPO baseline. \textit{Cost} is the cost reward, \textit{Completeness} is the fraction of orchestration decisions that are well formed and executable, and \textit{Steps Used} is the number of execution steps.}
\label{fig:training_dynamics}
\vspace{-2.0em}
\end{figure} 

\Cref{fig:agent_usage} shows that the optimized policy develops benchmark-dependent executor preferences.
Over training, the share of calls to the weaker and cheaper \texttt{model\_a} rises from $59\%$ to $75\%$ on MATH500 and from $62\%$ to $77\%$ on GAIA, but falls from $55\%$ to $32\%$ on ALFWorld in favor of the stronger \texttt{model\_b}, so the policy defaults to neither the cheapest nor the strongest executor and instead allocates according to task difficulty and the capability--cost profiles.
The corresponding training dynamics in \Cref{fig:training_dynamics} show \name maintaining higher efficiency and completeness than the standard-GRPO baseline while progressively reducing execution steps, ending at roughly $27$ steps against $36$ for the baseline and, after an initial dip, at higher validation performance.
These curves are consistent with the ablations, as tree-structured comparison and separately normalized rewards jointly favor trajectories that are better formed and less computationally redundant. 
Orchestration behavior cases can be found in \Cref{app:case_study}.

\section{Conclusion}

We introduced \name, which trains an LLM orchestrator with reinforcement learning over a milestone-plan-act workflow to adapt to evolving task states.
Tree-structured rollouts compare decisions under shared execution histories, with separately normalized rewards for task performance, execution cost, and planning completeness.
Across seven benchmarks, \name achieves the best task performance and higher intelligence efficiency, while benefiting from extension to larger executor pools and richer capability executors. 
Ablations and behavior analyses support the contributions of each component.

\clearpage
\bibliography{refs}
\bibliographystyle{template}

\clearpage
\appendix 
\part{Appendix}
\startcontents
\printcontents{}{1}{\textbf{Contents of Appendix}\vskip3pt\hrule\vskip5pt}
\vskip3pt\hrule\vskip5pt
\clearpage 

\section{Additional Implementation Details}
\label{app:implementation_details}

\subsection{Prompt for Milestone Generation} 
\label{app:milestone_prompt} 

Prompt~\ref{pt:milestone_prompt} gives the prompt used by the orchestrator $\pi_{\text{orch}}$ to generate the next milestone.

\begin{prompt}{Prompt for Next Milestone Generation}{milestone_prompt}
{\ttfamily\raggedright
\PromptLine{You are a strategic orchestrator driving a task to completion one milestone at a time.}
\PromptBlank
\PromptLine{=== Task ===}
\PromptLine{{task}}
\PromptBlank
\PromptLine{=== Milestones completed so far ===}
\PromptLine{{milestone_history}}
\PromptBlank
\PromptLine{=== Instructions ===}
\PromptLine{Examine the task and the results of every milestone completed so far, then decide between the following two options.}
\PromptBlank
\PromptLine{Option A. The task is NOT yet fully solved. Return:}
\PromptIndent{{"task_solved": false, "milestone": "<single imperative sentence describing the next concrete checkpoint>" }}
\PromptLine{Guidelines for the next milestone:}
\PromptLine{- It must be independently verifiable and build directly on what has already been done.}
\PromptLine{- Write it as a single, imperative sentence (e.g., "Verify that the purchased item matches the target specification").}
\PromptLine{- Do NOT describe sub-steps or implementation details.}
\PromptLine{- Do NOT add a milestone that merely restates the final answer. Synthesis is handled automatically.}
\PromptBlank
\PromptLine{Option B. The task IS fully solved based on the milestone results above. Return:}
\PromptIndent{{"task_solved": true, "milestone": ""}}
\PromptLine{Choose Option B only when the completed milestones collectively provide everything needed to answer the original task with confidence. Do not stop early.}
\PromptBlank
\PromptLine{Respond with a JSON object matching the required schema.}
}
\end{prompt}

\subsection{Prompt for Plan Generation} 
\label{app:plan_prompt}

Prompt~\ref{pt:plan_prompt} gives the prompt used by $\pi_{\text{orch}}$ to construct an execution plan for a selected milestone.

\begin{prompt}{Prompt for Plan Generation}{plan_prompt}
{\ttfamily\raggedright
\PromptLine{You are an expert task orchestrator responsible for producing an executable plan for a single milestone within a larger task.}
\PromptBlank
\PromptLine{=== Context ===}
\PromptLine{Overall task:}
\PromptIndent{{task}}
\PromptLine{Results from previously completed milestones:}
\PromptIndent{{previous_milestone_results}}
\PromptLine{Current milestone to accomplish:}
\PromptIndent{{current_milestone}}
\PromptBlank
\PromptLine{=== Available resources ===}
\PromptLine{Models (use EXACTLY one of these identifiers per step):}
\PromptIndent{{available_models}}
\PromptLine{Model intelligence ranking (weakest -> strongest):}
\PromptIndent{{model_intelligence_rank}}
\PromptLine{Model cost ranking (cheapest -> most expensive):}
\PromptIndent{{model_cost_rank}}
\PromptLine{Available tools:}
\PromptIndent{{available_tools}}
\PromptBlank
\PromptLine{=== Planning rules ===}
\PromptLine{1. Break the milestone into the minimal set of sub-tasks that together accomplish it.}
\PromptLine{2. Each step must have a unique integer "id" starting from 1.}
\PromptLine{3. A step's "dependencies" lists the ids of steps that must complete before it starts. Use an empty list [] for steps that can start immediately.}
\PromptLine{4. Assign the cheapest model that is sufficiently capable for each step. Reserve stronger (more expensive) models only for reasoning-heavy steps.}
\PromptLine{5. Keep sub-task descriptions concise and self-contained. The executor will receive only the sub-task text plus outputs from its dependencies.}
\PromptLine{6. Prefer parallelism: steps with no shared dependencies will run concurrently.}
\PromptLine{7. Do NOT add a "summarize results" step. That is handled automatically after execution.}
\PromptBlank
\PromptLine{=== Output ===}
\PromptLine{Respond with a JSON object matching the required schema. Each element of the "steps" list must contain:}
\PromptLine{- "id": integer}
\PromptLine{- "sub_task": string (what the executor must do)}
\PromptLine{- "model": string (exact model identifier from the list above)}
\PromptLine{- "dependencies": list of integer ids}
}
\end{prompt}

\subsection{Executor Prompt}

Prompt~\ref{pt:agent_pool_prompt} gives the prompt supplied to an executor $a_k\in\mathcal{A}$ for an assigned plan step. 

\begin{prompt}{Executor Prompt}{agent_pool_prompt}
{\ttfamily\raggedright
\PromptLine{You are a focused executor agent. Your job is to accomplish exactly one sub-task using the tools available to you.}
\PromptBlank
\PromptLine{=== Overall task (for context only) ===}
\PromptLine{{task}}
\PromptBlank
\PromptLine{=== Completed milestone results ===}
\PromptLine{{previous_milestone_results}}
\PromptBlank
\PromptLine{=== Outputs from upstream steps in the current plan ===}
\PromptLine{{previous_execution_logs}}
\PromptBlank
\PromptLine{=== Your sub-task ===}
\PromptLine{{sub_task}}
\PromptBlank
\PromptLine{=== Instructions ===}
\PromptLine{- Use tools as needed to gather information or perform actions.}
\PromptLine{- Think step-by-step, but keep intermediate reasoning brief.}
\PromptRaw{\detokenize{- When you have a confident answer, wrap it in }\Boxed{<your answer here>}\detokenize{ and stop. Do not add further commentary after the boxed answer.}}
\PromptRaw{\detokenize{- If you cannot complete the sub-task (e.g. a required resource is unavailable), state the reason clearly inside }\Boxed{FAILED: <reason>}\detokenize{.}}
\PromptLine{- Do not attempt to accomplish anything beyond the stated sub-task.}
}
\end{prompt}

\subsection{Prompt for Milestone Outcome Generation} 
\label{app:milestone_outcome_prompt}

Prompt~\ref{pt:milestone_outcome_prompt} gives the prompt used to aggregate executor outputs into a milestone-level outcome. 

\begin{prompt}{Prompt for Milestone Outcome Generation}{milestone_outcome_prompt}
{\ttfamily\raggedright
\PromptLine{You are a synthesis assistant. Given the outputs of several executor steps that together addressed a single milestone, produce a concise, accurate summary of what was accomplished.}
\PromptBlank
\PromptLine{=== Milestone ===}
\PromptLine{{current_milestone}}
\PromptBlank
\PromptLine{=== Step results ===}
\PromptLine{{plan_results}}
\PromptBlank
\PromptLine{=== Instructions ===}
\PromptLine{- Synthesize the step results into a single coherent answer for the milestone.}
\PromptLine{- Include all factual findings that are relevant to the milestone, omit execution details (tool calls, intermediate reasoning, errors) unless they materially affect the answer.}
\PromptLine{- If some steps failed or returned no useful output, note what is missing and reason from available information.}
\PromptRaw{\detokenize{- State your final milestone answer inside }\Boxed{<answer>}\detokenize{. The boxed content should be a standalone, readable summary (1-5 sentences or a concise structured list if appropriate).}}
}
\end{prompt}

\subsection{Prompt for Final Answer Generation} 

Prompt~\ref{pt:final_answer_prompt} gives the prompt used to synthesize the final answer from the completed milestone outcomes. 

\begin{prompt}{Prompt for Final Answer Generation}{final_answer_prompt}
{\ttfamily\raggedright
\PromptLine{You are a final-answer synthesizer. You have been given a task and the summarized results for each milestone that was executed to accomplish it. Your role is to produce the definitive answer to the original task.}
\PromptBlank
\PromptLine{=== Original task ===}
\PromptLine{{task}}
\PromptBlank
\PromptLine{=== Milestone results ===}
\PromptLine{{milestone_results}}
\PromptBlank
\PromptLine{=== Instructions ===}
\PromptLine{- Integrate all milestone results into a single, complete, and coherent answer to the original task.}
\PromptLine{- Present information at the level of detail the task requires: a factual lookup warrants a direct answer, an analytical task warrants structured reasoning.}
\PromptLine{- Do not expose internal workflow details (milestones, plan steps, tool usage).}
\PromptLine{- If any milestone failed or produced insufficient results, acknowledge the limitation concisely and answer as accurately as possible from available data.}
\PromptRaw{\detokenize{- Place your final answer inside }\Boxed{<answer>}\detokenize{. Everything outside the box is discarded.}}
}
\end{prompt}

\subsection{Prompt for CompassVerifier} 
\label{app:verifier_prompt}

Prompt~\ref{pt:verfifier_prompt} gives the evaluation prompt used by CompassVerifier~\citep{LiuLLXGLGZWZC25}. 

\begin{prompt}{Prompt for CompassVerifier}{verfifier_prompt}
{\ttfamily\raggedright
\PromptLine{Please as a grading expert, judge whether the final answers given by the candidates below are consistent with the standard answers, that is, whether the candidates answered correctly.}
\PromptBlank
\PromptLine{Here are some evaluation criteria:}
\PromptLine{1. Please refer to the given standard answer. You don't need to re-generate the answer to the question because the standard answer has been given. You only need to judge whether the candidate's answer is consistent with the standard answer according to the form of the question. THE STANDARD ANSWER IS ALWAYS CORRECT AND THE QUESTION IS PERFECTLY VALID. NEVER QUESTION THEM.}
\PromptLine{2. ONLY compare the FINAL ANSWER - COMPLETELY IGNORE any potential errors in the REASONING PROCESSES.}
\PromptLine{3. Some answers may be expressed in different ways, such as some answers may be a mathematical expression, some answers may be a textual description, as long as the meaning expressed is the same. Before making a judgment, please understand the question and the standard answer first, and then judge whether the candidate's answer is correct.}
\PromptLine{4. Some answers may consist of multiple items, such as multiple-choice questions, multiple-select questions, fill-in-the-blank questions, etc. Regardless of the question type, the final answer will be considered correct as long as it matches the standard answer, regardless of whether the reasoning process is correct. For multiple-select questions and multi-blank fill-in-the-blank questions, all corresponding options or blanks must be answered correctly and match the standard answer exactly to be deemed correct.}
\PromptRaw{\detokenize{5. If the prediction is given with }\Boxed{}\detokenize{, please ignore the }\Boxed{}\detokenize{ and only judge whether the candidate's answer is consistent with the standard answer.}}
\PromptLine{6. If the candidate's answer is invalid (e.g., incomplete (cut off mid-response), lots of unnormal repetitive content, or irrelevant to the question, saying it can't answer the question because some irresistible factors, like ethical issues, no enough information, etc.), select option C (INVALID).}
\PromptBlank
\PromptLine{Please judge whether the following answers are consistent with the standard answer based on the above criteria. Grade the predicted answer of this new question as one of:}
\PromptLine{A: CORRECT}
\PromptLine{B: INCORRECT}
\PromptLine{C: INVALID}
\PromptBlank
\PromptLine{Just return the letters "A", "B", or "C", with no text around it.}
\PromptBlank
\PromptLine{Here is your task. Simply reply with either CORRECT, INCORRECT, or INVALID. Don't apologize or correct yourself if there was a mistake, we are just trying to grade the answer.}
\PromptBlank
\PromptLine{<Original Question Begin>:}
\PromptLine{{question}}
\PromptLine{<Original Question End>}
\PromptBlank
\PromptLine{<Standard Answer Begin>:}
\PromptLine{{gold_answer}}
\PromptLine{<Standard Answer End>}
\PromptBlank
\PromptLine{<Candidate's Answer Begin>:}
\PromptLine{{llm_response}}
\PromptLine{<Candidate's Answer End>}
\PromptBlank
\PromptLine{Judging the correctness of the candidate's answer:}
}
\end{prompt}

\subsection{Executor Pool Configuration}
\label{app:configuration_agent_pool}

The following YAML example illustrates how we instantiate the executor pool $\mathcal{A}$ and expose its operational profiles to the orchestrator.
The orchestrator observes anonymous identifiers, such as \texttt{model\_a}, together with capability and cost rankings rather than recognizable model names.
The same interface can represent additional executors or profile attributes without changing the milestone-plan-act workflow.

\vspace{1.0em}
\begin{minted}[
    frame=single, 
    linenos=true, 
    breaklines, 
    fontsize=\small, 
    bgcolor=gray!10,      
    xleftmargin=2.5em    
]{yaml}
model_list:
  - model_a 
  - model_b 

model_names:
  model_a: Qwen/Qwen2.5-7B-Instruct
  model_b: Qwen/Qwen2.5-7B-Instruct

model_urls:
  model_a: xxxx
  model_b: xxxx

context_windows:
  model_a: 2048
  model_b: 4096

intelligence_rank:
  - model_a
  - model_b 

cost_rank:
  - model_a
  - model_b
\end{minted}
\vspace{1.0em} 

\subsection{Tool Configuration}
\label{app:tools}

\paragraph{Web Search.} Tool Card~\ref{tl:web_search_tool} specifies the web-search interface used in our experiments.
It accepts a query string and returns the top five results from the Bing Search API.
During training, we use ZeroSearch~\citep{abs-2505-04588} to simulate search results and reduce API cost.

\begin{tool}{Web Search}{web_search_tool} 
{\ttfamily\raggedright
\PromptLine{=== Name ===}
\PromptBlank
\PromptLine{WebSearch}
\PromptBlank
\PromptLine{=== Description ===}
\PromptBlank
\PromptLine{Retrieve the top five results for the given query through the web search.}
\PromptBlank
\PromptLine{=== Parameters ===} 
\PromptBlank
\begin{minted}[frame=single, breaklines, fontsize=\small]{json}
{
    "type": "object",
    "properties": {
        "query": {
            "type": "string",
            "description": "Query string. Include a single search query."
        }
    },
    "required": ["query"]
}
\end{minted}
}
\end{tool} 

\paragraph{Python Code Interpreter.} Tool Card~\ref{tl:code_interpreter} specifies the Python-execution interface.
We use SandboxFusion~\citep{sandbox_fusion} as the sandboxed execution backend.

\begin{tool}{Python Code Interpreter}{code_interpreter} 
{\ttfamily\raggedright
\PromptLine{=== Name ===}
\PromptBlank
\PromptLine{PythonInterpreter}
\PromptBlank
\PromptLine{=== Description ===}
\PromptBlank
\PromptLine{Execute Python code in a sandboxed environment. Use this to run Python code and get the execution results.}
\PromptBlank
\PromptLine{\textbf{Make sure to use print() for any output you want to see in the results.}}
\PromptBlank
\PromptLine{=== Parameters ===} 
\PromptBlank
\begin{minted}[frame=single, breaklines, fontsize=\small]{json}
{
    "type": "object",
    "properties": {
        "code": {
            "type": "string",
            "description": "The Python code to execute. Remember to use print() statements for any output you want to see."
        }
    },
    "required": ["code"]
}
\end{minted}
}
\end{tool} 

\paragraph{File Parser.} Tool Card~\ref{tl:file_parser} specifies the file-parsing interface, which accepts a file path and returns the parsed content.

\begin{tool}{File Parser}{file_parser} 
{\ttfamily\raggedright
\PromptLine{=== Name ===}
\PromptBlank
\PromptLine{FileParser}
\PromptBlank
\PromptLine{=== Description ===}
\PromptBlank
\PromptLine{Parse a file and return its content. Supports: text (.txt, .md, .py), spreadsheets (.xlsx, .csv), documents (.pdf, .docx, .pptx), data files (.json, .jsonl, .xml, .pdb), archives (.zip), and audio files with transcription (.mp3, .wav, .flac, .m4a, .ogg). }
\PromptBlank
\PromptLine{=== Parameters ===} 
\PromptBlank
\begin{minted}[frame=single, breaklines, fontsize=\small]{json}
{
    "type": "object",
    "properties": {
        "file_name": {
            "type": "string",
            "description": "Name of the file to parse (relative to the file base directory).",
            "examples": ["report.pdf", "data.csv", "audio.mp3"]
        }
    },
    "required": ["file_name"]
}
\end{minted}
}
\end{tool} 

\paragraph{Image Question Answering.} Tool Card~\ref{tl:image_qa} specifies the visual question-answering interface.
It accepts an image and a question and uses Qwen2.5-VL-32B-Instruct~\citep{abs-2502-13923} to produce the answer.

\begin{tool}{Image Question Answering}{image_qa}
{\ttfamily\raggedright
\PromptLine{=== Name ===}
\PromptBlank
\PromptLine{ImageQuestionAnswering}
\PromptBlank
\PromptLine{=== Description ===}
\PromptBlank
\PromptLine{Answer the question based on the image. Help with image understanding, object recognition, scene analysis, and answering questions about visual content. }
\PromptBlank
\PromptLine{=== Parameters ===}
\PromptBlank
\begin{minted}[frame=single, breaklines, fontsize=\small]{json}
{
    "type": "object",
    "properties": {
        "image_name": {
            "type": "string",
            "description": "Image name to be analyzed."
        },
        "question": {
            "type": "string",
            "description": "A clear, concise question about the image's content. For best results, ask straightforward factual questions rather than complex or multi-step reasoning questions."
        }
    },
    "required": ["image_name", "question"]
}
\end{minted}
}
\end{tool} 

\paragraph{Environment-Specific Tools for ALFWorld.} Tool Cards~\ref{tl:goto}--\ref{tl:heat} specify the interfaces used to interact with ALFWorld~\citep{ShridharYCBTH21}.

\begin{tool}{ALFWorld Goto}{goto} 
{\ttfamily\raggedright
\PromptLine{=== Name ===}
\PromptBlank
\PromptLine{goto}
\PromptBlank
\PromptLine{=== Description ===}
\PromptBlank
\PromptLine{Navigate to a specific receptacle or location in the environment. Move you to the target location and interact with objects there. You must be at the receptacle's location to interact with it. After successfully going to a location, 'You arrive at [location]' confirmation will be shown. }
\PromptBlank
\PromptLine{=== Parameters ===}
\PromptBlank
\begin{minted}[frame=single, breaklines, fontsize=\small]{json}
{
    "type": "object",
    "properties": {
        "recep": {
            "type": "string",
            "description": "Name of the target receptacle or location (e.g., 'cabinet 1', 'fridge 1', 'countertop 2')."
        }
    },
    "required": ["recep"]
}
\end{minted}
}
\end{tool} 

\begin{tool}{ALFWorld Open}{open} 
{\ttfamily\raggedright
\PromptLine{=== Name ===}
\PromptBlank
\PromptLine{open}
\PromptBlank
\PromptLine{=== Description ===}
\PromptBlank
\PromptLine{Open a closed receptacle to access its contents. Is necessary before you can see or interact with objects inside containers like cabinets, drawers, fridge, microwave, or safe. You must be at the receptacle's location to open it. After successfully opening a receptacle, 'The [receptacle] is open' confirmation will be shown.}
\PromptBlank
\PromptLine{=== Parameters ===}
\PromptBlank
\begin{minted}[frame=single, breaklines, fontsize=\small]{json}
{
    "type": "object",
    "properties": {
        "recep": {
            "type": "string",
            "description": "Name of the receptacle to open (e.g., 'cabinet 2', 'fridge 1', 'microwave 1')."
        }
    },
    "required": ["recep"]
}
\end{minted}
}
\end{tool} 

\begin{tool}{ALFWorld Use}{use} 
{\ttfamily\raggedright
\PromptLine{=== Name ===}
\PromptBlank
\PromptLine{use}
\PromptBlank
\PromptLine{=== Description ===}
\PromptBlank
\PromptLine{Use an object or receptacle. You must be at the object's location to use it. The commonly used object is desklamps for lighting objects. After successfully using an desklamp, 'You turn on [object]' confirmation will be shown.}
\PromptBlank
\PromptLine{=== Parameters ===}
\PromptBlank
\begin{minted}[frame=single, breaklines, fontsize=\small]{json}
{
    "type": "object",
    "properties": {
        "obj": {
            "type": "string",
            "description": "Name of the object or receptacle to use (e.g., 'desklamp 1')."
        }
    },
    "required": ["obj"]
}
\end{minted}
}
\end{tool} 

\begin{tool}{ALFWorld Take}{take} 
{\ttfamily\raggedright
\PromptBlank
\PromptLine{=== Name ===}
\PromptBlank
\PromptLine{take}
\PromptBlank
\PromptLine{=== Description ===} 
\PromptBlank
\PromptLine{Pick up an object from a receptacle or surface. Grab items like food, utensils, containers, and other objects from their current location. Make sure the receptacle is open if the object is inside (cabinets, drawers, fridge, microwave, safe). Only hold one object at a time - must move the current object before taking another. After successfully taking an object, 'You pick up [object]' confirmation will be shown.}
\PromptBlank
\PromptLine{=== Parameters ===} 
\PromptBlank
\begin{minted}[frame=single, breaklines, fontsize=\small]{json}
{
    "type": "object",
    "properties": {
        "obj": {
            "type": "string",
            "description": "Name of the object to pick up (e.g., 'apple 1', 'mug 2', 'knife 1')."
        },
        "from": {
            "type": "string",
            "description": "Name of the source receptacle or surface (e.g., 'countertop 1', 'cabinet 4')."
        }
    },
    "required": ["obj", "from"]
}
\end{minted}
}
\end{tool} 

\begin{tool}{ALFWorld Clean}{clean} 
{\ttfamily\raggedright
\PromptLine{=== Name ===}
\PromptBlank
\PromptLine{clean}
\PromptBlank
\PromptLine{=== Description ===} 
\PromptBlank
\PromptLine{Clean an object using a cleaning receptacle. Wash or clean items like dishes, utensils, or food items. The most common cleaning receptacle is the sinkbasin. Must have the object in the inventory first (use 'take'). After successfully cleaning an object, 'You clean [object] using [receptacle]' confirmation will be shown.}
\PromptBlank
\PromptLine{=== Parameters ===} 
\PromptBlank
\begin{minted}[frame=single, breaklines, fontsize=\small]{json}
{
    "type": "object",
    "properties": {
        "obj": {
            "type": "string",
            "description": "Name of the object to clean (e.g., 'mug 1', 'knife 1')."
        },
        "with": {
            "type": "string",
            "description": "Receptacle used for cleaning (e.g., 'sinkbasin 1')."
        }
    },
    "required": ["obj", "with"]
}
\end{minted}
}
\end{tool} 

\begin{tool}{ALFWorld Heat}{heat} 
{\ttfamily\raggedright
\PromptLine{=== Name ===}
\PromptBlank
\PromptLine{heat}
\PromptBlank
\PromptLine{=== Description ===} 

\PromptLine{Heat an object using a heating receptacle. Warm up items like food or containers to make them hot. Must have the object in the inventory first (use 'take'). The microwave is the primary heating location. If get 'Nothing happens', make sure at the microwave location and have the object in the inventory. After successfully heating an object, 'You heat [object] using [receptacle]' confirmation will be shown.}
\PromptBlank
\PromptLine{=== Parameters ===} 
\PromptBlank
\begin{minted}[frame=single, breaklines, fontsize=\small]{json}
{
    "type": "object",
    "properties": {
        "obj": {
            "type": "string",
            "description": "Name of the object to heat (e.g., 'apple 1', 'mug 1')."
        },
        "with": {
            "type": "string",
            "description": "Receptacle used for heating (e.g., 'microwave 1')."
        }
    },
    "required": ["obj", "with"]
}
\end{minted}
}
\end{tool} 

\paragraph{Environment-Specific Tools for WebShop.} Tool Cards~\ref{tl:webshop_search}--\ref{tl:webshop_click} specify the interfaces used to interact with WebShop~\citep{Yao0YN22}.

\begin{tool}{WebShop Search}{webshop_search} 
{\ttfamily\raggedright
\PromptLine{=== Name ===}
\PromptBlank
\PromptLine{search}
\PromptBlank
\PromptLine{=== Description ===} 
\PromptBlank
\PromptLine{Search for products with the given query. Use this only if a [Search] button appears in the tool call result. Note: Click the [Back to Search] button if there's no [Search] button via the `click` tool, and then try again with a more specific query.}
\PromptBlank
\PromptLine{=== Parameters ===} 
\PromptBlank
\begin{minted}[frame=single, breaklines, fontsize=\small]{json}
{
    "type": "object",
    "properties": {
        "query": {
            "type": "string",
            "description": "Search query for products"
        }
    },
    "required": ["query"]
}
\end{minted}
|}
\end{tool} 

\begin{tool}{WebShop Click}{webshop_click} 
{\ttfamily\raggedright
\PromptLine{=== Name ===}
\PromptBlank
\PromptLine{click}
\PromptBlank
\PromptLine{=== Description ===} 
\PromptBlank
\PromptLine{Click on a button or product. Use this only if a [button] is present in the tool call result. Click the [Buy Now] button on its product page to finish the shopping task if the most suitable product is identified.}
\PromptBlank
\PromptLine{=== Parameters ===} 
\PromptBlank
\begin{minted}[frame=single, breaklines, fontsize=\small]{json}
{
    "type": "object",
    "properties": {
        "button": {
            "type": "string",
            "description": "Name of the button or product ASIN, don't add '[]' around the button name"
        }
    },
    "required": ["button"]
}
\end{minted}
}
\end{tool}

\subsection{Dataset Details}

We summarize the data sources used for training and evaluation in this section.

\subsubsection{Training Data}

\paragraph{DAPO-Math.}
DAPO-Math~\citep{abs-2503-14476} is a curated mathematical-reasoning dataset introduced with the DAPO reinforcement-learning system.
Its commonly used DAPO-Math-17K split contains 17K competition-style prompts with integer answers, supporting verifiable answer-level rewards without requiring reference solution traces.

\paragraph{OpenScienceReasoning-2.}
OpenScienceReasoning-2~\citep{open_science_reasoning_2} contains synthetic reasoning examples across scientific and academic domains.
We include it to expose the orchestrator to knowledge-intensive tasks beyond mathematical problem solving.

\paragraph{Natural Questions.}
Natural Questions~\citep{KwiatkowskiPRCP19} contains more than 300K training examples constructed from real information-seeking queries, with long- and short-answer annotations derived from Wikipedia pages.
We use it to train retrieval-oriented orchestration and evidence integration.

\paragraph{HotpotQA.}
HotpotQA~\citep{Yang0ZBCSM18} contains approximately 113K multi-hop questions with supporting-fact annotations across multiple Wikipedia documents.
It provides training examples for decomposing evidence-gathering tasks and integrating information across sources.

\paragraph{WebDancer.}
WebDancer~\citep{abs-2505-22648} provides data for agentic information seeking that involves iterative browsing, evidence collection, and answer synthesis.
We use these examples to train long-horizon web-research behavior.

\paragraph{ALFWorld and WebShop.}
We additionally include instances from ALFWorld~\citep{ShridharYCBTH21} and WebShop~\citep{Yao0YN22} to train sequential environment interaction and tool-mediated decision making. 

\subsubsection{Evaluation Benchmarks}

\paragraph{AIME24 and AIME25.}
AIME24 and AIME25 comprise problems from the 2024 and 2025 American Invitational Mathematics Examinations.
Each problem has a three-digit integer answer in the range 000--999 and requires multi-step competition-level reasoning, enabling automatic answer verification.

\paragraph{MATH500.}
MATH500~\citep{HendrycksBKABTS21} is a 500-problem subset of MATH spanning algebra, number theory, geometry, counting and probability, prealgebra, and precalculus.
We use it to evaluate mathematical reasoning across a broader range of topics and difficulty levels than AIME.

\paragraph{ALFWorld.}
ALFWorld~\citep{ShridharYCBTH21} is a text-based interactive benchmark aligned with the embodied ALFRED environment.
It covers six household task types requiring long-horizon planning, environment grounding, and execution through actions such as navigation, object manipulation, heating, cooling, and cleaning.

\paragraph{WebShop.}
WebShop~\citep{Yao0YN22} is a simulated e-commerce environment in which an agent must satisfy a natural-language shopping request through search, product inspection, attribute comparison, and purchase actions.
It evaluates grounded sequential decision making under user-specified constraints.

\paragraph{GAIA.}
GAIA~\citep{MialonF0LS24} evaluates general assistants on real-world questions requiring combinations of reasoning, browsing, tool use, file manipulation, and multimodal understanding.
We use it to assess long-horizon information gathering and evidence integration.

\paragraph{HLE.}
Humanity's Last Exam (HLE)~\citep{abs-2501-14249} is a challenging multimodal benchmark spanning mathematics, science, the humanities, and other specialized domains.
Its expert-written, closed-ended questions evaluate broad-domain knowledge and precise answer generation.

\subsection{Baseline Details}\label{app:baseline_details}

We group the evaluated methods into single-agent approaches, multi-agent collaboration frameworks, and agentic orchestration methods. Unless otherwise stated, the baselines use the common executor backbones and tool interfaces described in the main text.

\paragraph{\textsc{Direct Reasoning}.}
This baseline prompts a single executor to solve each task directly. 
It does not use iterative environment interaction, multi-agent collaboration, or a learned orchestration policy. 
In contrast, \name explicitly orchestrates multiple executors through iterative milestone decomposition, dependency-aware planning, and adaptive executor assignment.

\paragraph{\textsc{ReAct}.}
\textsc{ReAct}~\citep{YaoZYDSN023} interleaves natural-language reasoning, actions, and environment observations. We use it as an interactive single-agent baseline with access to the same task-specific tools, but without explicit coordination among multiple executors. 
Unlike \textsc{ReAct}, which makes reasoning and tool-use decisions within a single-agent trajectory, \name additionally learns how to decompose tasks across milestones, construct dependency-aware execution plans, and allocate subtasks among heterogeneous executors.

\paragraph{\textsc{Reflexion}.}
\textsc{Reflexion}~\citep{ShinnCGNY23} uses evaluator feedback to produce verbal reflections that condition subsequent attempts. It provides inference-time self-improvement through episodic memory without updating model parameters. 
In contrast, \name uses execution feedback during reinforcement learning to optimize an explicit orchestration policy, with feedback assigned to intermediate milestone and plan decisions rather than used solely as textual context for subsequent attempts.

\paragraph{\textsc{AutoGen}.}
\textsc{AutoGen}~\citep{abs-2308-08155} supports task solving through role-based multi-agent conversations. We use its default two-agent configuration, in which an assistant performs reasoning and response generation while a user-proxy agent mediates requests and, when needed, tool or code execution. 
Whereas \textsc{AutoGen} coordinates agents through a predefined conversational structure, \name learns an execution-aware orchestration policy that dynamically determines milestones, subtask dependencies, and executor assignments as execution progresses.

\paragraph{\textsc{Multi-Agent Debate}.}
Multi-Agent Debate~\citep{Du00TM24} elicits candidate solutions from multiple agents, allows them to exchange critiques, and aggregates their reasoning into a final answer. The interaction follows a fixed debate protocol rather than a profile-conditioned runtime orchestration policy. 
In contrast, \name does not prescribe a fixed communication pattern. 
It dynamically constructs dependency-aware execution graphs and delegates different subtasks according to the current execution history and executor profiles.

\paragraph{\textsc{MAS-ZERO}.}
\textsc{MAS-ZERO}~\citep{abs-2505-14996} is a training-free method that constructs a problem-specific multi-agent configuration at inference time. A meta-agent decomposes the task, critiques intermediate solutions, and refines the agent composition and workflow without relying on a validation set. 
Both \textsc{MAS-ZERO} and \name adapt multi-agent structures to individual tasks, but they differ in how orchestration is obtained: \textsc{MAS-ZERO} performs training-free inference-time construction, whereas \name learns an orchestration policy from execution feedback and organizes execution through iterative milestone-level plans with explicit decision-level credit assignment.

\paragraph{\textsc{DAAO}.}
Difficulty-Aware Agentic Orchestration (\textsc{DAAO})~\citep{SuLXSTSH26} estimates query difficulty and adapts workflow depth, operators, and model assignments accordingly. It provides query-specific workflow construction through a difficulty-aware inference-time controller. 
In contrast, \name conditions orchestration not only on the initial query but also on accumulated execution outcomes, allowing subsequent milestones and plans to evolve with the task state. 
Its orchestration policy is further optimized through reinforcement learning rather than determined solely by an inference-time controller.

\paragraph{\textsc{AgentFlow}.}
\textsc{AgentFlow}~\citep{abs-2510-05592} coordinates a planner, executor, verifier, and generator through shared evolving memory. It trains the planner within this iterative interaction loop using trajectory-level feedback. 
Similar to \name, \textsc{AgentFlow} learns orchestration through reinforcement learning and supports iterative execution. 
The key difference is the optimization granularity: \name explicitly structures orchestration into milestone and plan decisions and constructs tree-structured rollouts from shared execution histories, enabling decision-level performance and cost feedback rather than assigning the same trajectory-level signal across the entire interaction.

\paragraph{\textsc{Puppeteer}.}
\textsc{Puppeteer}~\citep{abs-2505-19591} trains a centralized orchestrator to sequence and prioritize agents as the task state evolves. The orchestration policy is optimized from execution feedback rather than being specified entirely through a fixed prompt or workflow. 
Both \textsc{Puppeteer} and \name therefore learn adaptive orchestration policies, but \name additionally introduces the milestone-plan-act abstraction, represents each milestone plan as an explicit dependency graph that supports parallel execution, and performs decision-level optimization through tree-structured rollouts and multi-objective rewards.

\paragraph{\textsc{MAS-Orchestra}.}
\textsc{MAS-Orchestra}~\citep{abs-2601-14652} formulates holistic multi-agent construction as a function-calling reinforcement-learning problem. 
Its orchestrator jointly determines task decomposition, agent selection, dependencies, and information flow in a single orchestration decision. 
\name optimizes a closely related set of orchestration dimensions but distributes them across iterative milestone and plan decisions: each milestone locally scopes decomposition and dependency-aware execution, while subsequent milestones are conditioned on intermediate outcomes. 
This iterative abstraction further provides intermediate decision points for tree-structured credit assignment.

\paragraph{\textsc{MasRouter}.}
\textsc{MasRouter}~\citep{YueZLWWCQ25} uses cascaded controllers to determine the collaboration mode, allocate agent roles, and assign an LLM to each role. 
It therefore adapts both the multi-agent configuration and the underlying model allocation to each query. 
In contrast, \name treats executor selection as one component of a broader orchestration problem that also includes iterative milestone decomposition, dependency-aware planning, and execution scheduling, and learns these decisions jointly through a reinforcement-learned LLM orchestrator conditioned on executor profiles.

\subsection{Baselines Implementation Details} \label{app:baselines_implementation} 

For single-agent baselines, we instantiate the workflow with a single executor, \texttt{model\_b}.
For multi-agent collaboration baselines, we instantiate it with all executors in $\mathcal{A}$.
For the agentic orchestration baselines, we replace only the method's own model pool with $\mathcal{A}$, leaving its remaining components, such as its role pool, unchanged. 
For all baselines, we ensure the use of the same orchestrator backbones and tool interfaces as in \name.

\subsection{Training Parameters}

\Cref{tab:train_param} summarizes the hyperparameters used to optimize the orchestrator in \name.

\begin{table}[ht]
    \centering
    \caption{Training hyperparameters for the orchestrator.} \label{tab:train_param}
    \begin{tabular}{lc}
        \toprule
        \textbf{Parameter} & \textbf{Value} \\
        \midrule
        Batch Size & 64 \\
        Initial Trajectories ($B$) & 2 \\
        Branches per Expansion ($N$) & 4 \\
        Expansion Rounds ($E$) & 2 \\
        Rollout Temperature & 0.7 \\
        Rollout Top-\(p\) & 0.8 \\
        Rollout Top-\(k\) & 20 \\
        Maximum Number of Generation Tokens & 2048 \\
        Learning Rate & $1 \times 10^{-6}$ \\
        KL Loss Coefficient & $1 \times 10^{-4}$ \\
        $\epsilon_{l}$ & 0.2 \\
        $\epsilon_{r}$ & 0.28 \\
        Gradient Clipping & 1.0 \\
        Number of Training Steps & 150 \\ 
        $\lambda_{\mathrm{perf}}$ & 1.0 \\
        $\lambda_{\mathrm{cost}}$ & 0.1 \\
        $\lambda_{\mathrm{comp}}$ & 0.1 \\
        \bottomrule
    \end{tabular}
\end{table} 

\subsection{Definition of Intelligence Efficiency}
\label{app:intelligence_efficiency}

\paragraph{Cost of a Single Run.}
For one evaluation instance $q$, let $t_k(q)$ denote the number of output tokens generated by executor $a_k$ over all subtasks delegated to it while solving $q$, summed across every milestone and plan of the trajectory.
We define the \emph{cost-weighted executor output tokens} of that instance as
\begin{equation}
    T(q) = \sum_{k=1}^{K} \kappa_k \, t_k(q),
    \label{eq:weighted_tokens}
\end{equation}
where $\kappa_k$ is the relative cost of executor $a_k$ declared in its operational profile $\phi_k$ (\Cref{sec:preliminary}).
We set $\kappa_k$ to the executor's relative configured cost normalized by that of the cheapest executor in the pool, so that the cheapest executor has $\kappa_k=1$ and $T(q)$ is expressed in units of tokens of the cheapest executor.
In our two-executor configuration this gives $\kappa=(1.0,\,2.0)$ for \texttt{model\_a} and \texttt{model\_b}, and $(1.0,\,2.0,\,3.0)$ in the three-executor configuration of \Cref{app:general_executor}.
We report $T$ in thousands of tokens (K) and, unless stated otherwise, averaged over evaluation instances and over the repeated runs.

\paragraph{Intelligence Efficiency.}
Let $P$ denote the task performance of a method on a benchmark, in the units reported in \Cref{tab:math_results}, and let $T$ denote its mean cost-weighted executor output tokens per instance.
Intelligence efficiency is their ratio,
\begin{equation}
    \eta = \frac{P}{T},
    \label{eq:intelligence_efficiency}
\end{equation}
measured in performance points per 1K cost-weighted executor output tokens.
It answers how much task-solving capability a method extracts from a fixed execution budget, rather than how much capability it reaches at unbounded budget.
Its reciprocal $1/\eta = T/P$ is the expected cost-weighted token spend per \emph{solved} instance, which is the quantity we report when comparing token spend at matched capability.
The two summarize the same measurement, and we use whichever is more direct for the claim being made. 

\paragraph{Relation to the Cost Reward.}
The quantity $C(\nu)$ optimized by the cost reward in \Cref{eq:efficiency_reward} is the same cost-weighted output-token count of \Cref{eq:weighted_tokens}, restricted to the subtree rooted at decision $\nu$.
The training signal is a within-sibling $z$-score of $C(\nu)$, which is comparable only among alternatives sharing an execution history, whereas $\eta$ is an evaluation-time aggregate over complete trajectories.
Training therefore optimizes the numerator and denominator of $\eta$ through separate rewards, and $\eta$ is used only for reporting.

\section{Additional Experimental Results and Analysis}

\subsection{Scaling the Orchestrator}\label{app:orchestrator_scaling}

\Cref{tab:scaling_orchestrator_results} compares Qwen2.5-3B-Instruct and Qwen2.5-7B-Instruct as orchestrator backbones for \name.
Scaling the orchestrator from 3B to 7B improves performance on every reported benchmark, which suggests that the learned orchestration policy benefits from a stronger backbone.

\begin{table}[hbt]
\centering 
\caption{Performance of \name with 3B and 7B orchestrator backbones.}
\label{tab:scaling_orchestrator_results}
\resizebox{.97\linewidth}{!}{
\begin{tabular}{lccccccc}
\toprule
\multirow{2}{*}{\textbf{Method}} & \multirow{2}{*}{\textbf{Orchestrator Size}} & \multicolumn{2}{c}{\cellcolor{cyan!20} \textbf{Mathematical}} & \multicolumn{2}{c}{\cellcolor{Orchid!20} \textbf{Embodied}} & \multicolumn{2}{c}{\cellcolor{LimeGreen!20} \textbf{Deep Research}} \\
\cmidrule(lr){3-4} \cmidrule(lr){5-6} \cmidrule(lr){7-8}
& & \textbf{AIME24/25} & \textbf{MATH500} & \textbf{ALFWorld} & \textbf{WebShop} & \textbf{GAIA} & \textbf{HLE} \\
\midrule
\name & 3B & $32.3_{\pm 0.5}$/$28.7_{\pm 0.4}$ & $85.4_{\pm 0.4}$ & $66.7_{\pm 0.5}$ & $58.0_{\pm 0.6}$ & $15.2_{\pm 0.4}$ & $5.1_{\pm 0.3}$ \\
\name & 7B & $35.0_{\pm 0.4}$/$31.5_{\pm 0.5}$ & $90.4_{\pm 0.4}$ & $70.2_{\pm 0.5}$ & $62.7_{\pm 0.6}$ & $20.5_{\pm 0.4}$ & $7.8_{\pm 0.3}$ \\
\bottomrule
\end{tabular}
} 
\vspace{-0.8em}
\end{table} 

\subsection{Wall-Clock Time Analysis}  
\label{app:wall_clock_time}

\begin{figure}
    \centering
    \includegraphics[width=.95\linewidth]{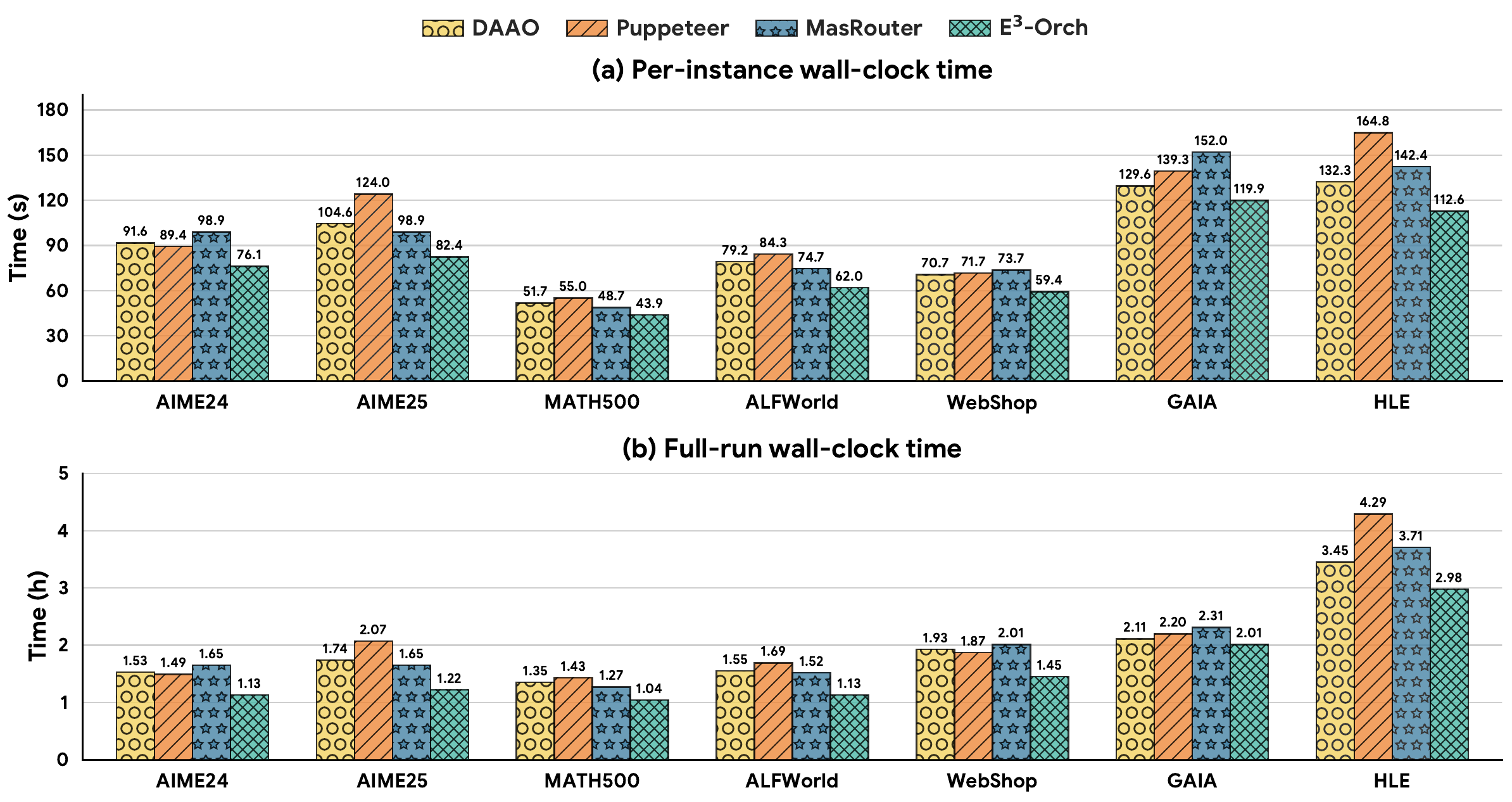}
    \vspace{-1.0em}
    \caption{Wall-clock time comparison of \name and baseline methods on evaluation benchmarks.}
    \label{fig:wall_clock_time}
\end{figure}

\paragraph{Evaluation Wall-Clock Time. }
To demonstrate the efficiency of \name, we also measure its wall-clock time on evaluation benchmarks and compare it against the baseline methods. 
As shown in \Cref{fig:wall_clock_time}, \name consistently achieves lower wall-clock times compared to the baseline methods across all evaluation benchmarks regarding per-instance and full-benchmark measurements. 

\begin{wraptable}{r}{.4\linewidth}
    \vspace{-1.0em}
    \centering
    \small
    \caption{Training wall-clock time of \name and the strongest trained baseline.}
    \label{tab:training_wall_clock_time}
    \begin{tabular}{@{}lc@{}}
        \toprule
        \bf Method & \bf Time (h) \\
        \midrule
        \name & 36.2 \\
        \textsc{AgentFlow} & 40.6 \\
        \bottomrule
    \end{tabular}
    \vspace{-1.0em}
\end{wraptable}
\paragraph{Training Wall-Clock Time.}
We further compare the trained methods by the wall-clock time required to complete reinforcement learning, measured end to end under identical hardware and software settings.
Tree-structured rollouts expand a branch from an existing decision node rather than re-executing a trajectory from scratch, so each additional candidate costs only the execution below its branch point.
As shown in \Cref{tab:training_wall_clock_time}, \name therefore completes training faster than \textsc{AgentFlow}, which performs a full rollout for every training instance.
Router-based methods such as \textsc{MasRouter} remain cheaper to train, since they do not require the RLVR process, but they do not match \name on task performance or on robustness to changes in the executor pool (\Cref{app:general_executor}).

\subsection{Generalization Across Executor Configurations}
\label{app:general_executor}

We evaluate whether the profile-conditioned interface transfers to executor configurations that differ from the two-executor setup in \Cref{app:configuration_agent_pool}.
Both tables report task performance and token cost, their ratio is the intelligence efficiency of \Cref{fig:efficiency}.

\paragraph{Generalization to Unseen Executor Configurations.}
We first expand the pool to three executors, adding a larger Qwen2.5-14B-Instruct as \texttt{model\_c}.
A larger pool is not free: the added executor is the most capable but also the most expensive one, so a policy that simply reaches for it converts the extra capability into extra cost.
\Cref{tab:model_list_genel} shows that \name does not.
It attains the best task performance on all seven benchmarks while also spending the fewest tokens, $12$--$37\%$ below the cheapest baseline in each column, for $1.46$--$2.92\times$ the intelligence efficiency of the stronger baseline.
\textsc{MasRouter} shows the opposite pattern: its cascaded controllers must now route over a larger action space, and its performance falls well below its two-executor result while its token cost stays above \name's on every benchmark.
Extending the pool therefore tests not only whether an orchestrator still functions, but whether it can decline the more expensive option when the task does not need it.

\vspace{1.0em}
\begin{minted}[
    frame=single, 
    linenos=true, 
    breaklines, 
    fontsize=\small, 
    bgcolor=gray!10,      
    xleftmargin=2.5em    
]{yaml}
model_list:
  - model_a 
  - model_b 
  - model_c

model_names:
  model_a: Qwen/Qwen2.5-7B-Instruct
  model_b: Qwen/Qwen2.5-7B-Instruct
  model_c: Qwen/Qwen2.5-14B-Instruct

model_urls:
  model_a: xxxx
  model_b: xxxx
  model_c: xxxx

context_windows:
  model_a: 512
  model_b: 2048
  model_c: 2048

intelligence_rank:
  - model_a
  - model_b 
  - model_c

cost_rank:
  - model_a
  - model_b
  - model_c
\end{minted}
\vspace{1.0em} 

\begin{table}[hbt]
    \centering 
    \caption{Task performance ($\uparrow$) and token cost ($\downarrow$, weighted executor output tokens per instance in thousands) under a three-executor configuration.}
    \label{tab:model_list_genel}
    \resizebox{.97\linewidth}{!}{
    \begin{tabular}{lccccccc}
    \toprule
    \multirow{2}{*}{\textbf{Method}} & \multirow{2}{*}{\textbf{Metric}}  & \multicolumn{2}{c}{\cellcolor{cyan!20} \textbf{Mathematical}} & \multicolumn{2}{c}{\cellcolor{Orchid!20} \textbf{Embodied}} & \multicolumn{2}{c}{\cellcolor{LimeGreen!20} \textbf{Deep Research}} \\
    \cmidrule(lr){3-4} \cmidrule(lr){5-6} \cmidrule(lr){7-8}
    & & \textbf{AIME24/25} & \textbf{MATH500} & \textbf{ALFWorld} & \textbf{WebShop} & \textbf{GAIA} & \textbf{HLE} \\
    \midrule
    \multirow{2}{*}{\textsc{AgentFlow}} & Performance & $34.0_{\pm 0.5}$/$28.8_{\pm 0.4}$ & $89.1_{\pm 0.4}$ & $66.5_{\pm 0.5}$ & $58.5_{\pm 0.6}$ & $17.4_{\pm 0.5}$ & $5.9_{\pm 0.4}$ \\
    & Tokens (K) & $11.8_{\pm 1.0}$/$12.9_{\pm 1.3}$ & $5.8_{\pm 0.6}$ & $4.7_{\pm 0.4}$ & $5.0_{\pm 0.5}$ & $23.6_{\pm 5.1}$ & $15.3_{\pm 2.0}$ \\
    \hdashline
    \multirow{2}{*}{\textsc{MasRouter}} & Performance & $20.0_{\pm 0.4}$/$16.8_{\pm 0.3}$ & $83.5_{\pm 0.5}$ & $52.0_{\pm 0.6}$ & $40.4_{\pm 0.4}$ & $8.5_{\pm 0.3}$ & $3.7_{\pm 0.3}$ \\
    & Tokens (K) & $8.8_{\pm 0.9}$/$12.3_{\pm 1.1}$ & $5.2_{\pm 0.7}$ & $5.0_{\pm 0.5}$ & $5.8_{\pm 0.5}$ & $15.3_{\pm 2.0}$ & $12.9_{\pm 1.3}$ \\
    \hdashline
    \multirow{2}{*}{\name} & Performance & \textbf{37.2}$_{\pm \mathbf{0.4}}$/\textbf{33.5}$_{\pm \mathbf{0.5}}$ & \textbf{92.1}$_{\pm \mathbf{0.3}}$ & \textbf{73.3}$_{\pm \mathbf{0.5}}$ & \textbf{64.9}$_{\pm \mathbf{0.4}}$ & \textbf{23.1}$_{\pm \mathbf{0.6}}$ & \textbf{9.6}$_{\pm \mathbf{0.3}}$ \\
    & Tokens (K) & \textbf{7.8}$_{\pm \mathbf{0.2}}$/\textbf{8.5}$_{\pm \mathbf{0.5}}$ & \textbf{4.0}$_{\pm \mathbf{0.4}}$ & \textbf{3.0}$_{\pm \mathbf{0.3}}$ & \textbf{3.3}$_{\pm \mathbf{0.4}}$ & \textbf{10.4}$_{\pm \mathbf{0.8}}$ & \textbf{10.4}$_{\pm \mathbf{0.8}}$ \\
    \bottomrule
    \end{tabular}
    } 
    \vspace{-0.8em}
\end{table}

\paragraph{Generalization to Enriched Executor Descriptions.}
We next enrich each executor profile with rankings of domain-specific capabilities, replacing \texttt{model\_b} with a mathematics-specialized executor and exposing separate general and mathematical intelligence rankings alongside the cost ranking.
This changes the schema the orchestrator reads, not just the pool it reads about.
In \Cref{tab:model_desc_genel}, \name again attains the best task performance and the lowest token cost on all seven benchmarks, spending $17$--$40\%$ fewer tokens than the cheapest baseline and reaching $1.61$--$2.63\times$ its intelligence efficiency, without any retraining for the revised schema.
Its margin over the baselines does not shrink under either extension, which is what the profile-conditioned interface is intended to provide: the orchestrator reads capability and cost from $\Phi$ rather than from a fixed executor set, so a richer or larger pool changes its input rather than its interface.

\vspace{1.0em}
\begin{minted}[
    frame=single, 
    linenos=true, 
    breaklines, 
    fontsize=\small, 
    bgcolor=gray!10,      
    xleftmargin=2.5em    
]{yaml}
model_list:
  - model_a 
  - model_b

model_names:
  model_a: Qwen/Qwen2.5-7B-Instruct
  model_b: Qwen/Qwen2.5-Math-7B-Instruct

model_urls:
  model_a: xxxx
  model_b: xxxx

context_windows:
  model_a: 2048
  model_b: 2048

general_intelligence_rank:
  - model_b
  - model_a

mathematical_intelligence_rank:
  - model_a
  - model_b

cost_rank:
  - model_a
  - model_b

\end{minted}
\vspace{1.0em} 

\begin{table}[hbt]
    \centering 
    \caption{Task performance ($\uparrow$) and token cost ($\downarrow$, weighted executor output tokens per instance in thousands) with enriched, domain-specific executor profiles.}
    \label{tab:model_desc_genel}
    \resizebox{.97\linewidth}{!}{
    \begin{tabular}{lccccccc}
    \toprule
    \multirow{2}{*}{\textbf{Method}} & \multirow{2}{*}{\textbf{Metric}}  & \multicolumn{2}{c}{\cellcolor{cyan!20} \textbf{Mathematical}} & \multicolumn{2}{c}{\cellcolor{Orchid!20} \textbf{Embodied}} & \multicolumn{2}{c}{\cellcolor{LimeGreen!20} \textbf{Deep Research}} \\
    \cmidrule(lr){3-4} \cmidrule(lr){5-6} \cmidrule(lr){7-8}
    & & \textbf{AIME24/25} & \textbf{MATH500} & \textbf{ALFWorld} & \textbf{WebShop} & \textbf{GAIA} & \textbf{HLE} \\
    \midrule
    \multirow{2}{*}{\textsc{AgentFlow}} & Performance & $32.8_{\pm 0.4}$/$27.0_{\pm 0.5}$ & $87.9_{\pm 0.5}$ & $65.6_{\pm 0.4}$ & $57.4_{\pm 0.6}$ & $15.6_{\pm 0.5}$ & $5.1_{\pm 0.3}$ \\
    & Tokens (K) & $12.3_{\pm 1.1}$/$12.3_{\pm 1.1}$ & $5.8_{\pm 0.6}$ & $4.8_{\pm 0.4}$ & $5.1_{\pm 0.6}$ & $20.0_{\pm 2.6}$ & $15.3_{\pm 2.0}$ \\
    \hdashline
    \multirow{2}{*}{\textsc{MasRouter}} & Performance & $18.8_{\pm 0.5}$/$15.1_{\pm 0.3}$ & $82.7_{\pm 0.6}$ & $51.2_{\pm 0.5}$ & $38.8_{\pm 0.4}$ & $7.7_{\pm 0.4}$ & $2.7_{\pm 0.3}$ \\
    & Tokens (K) & $9.7_{\pm 1.0}$/$11.8_{\pm 1.0}$ & $5.5_{\pm 0.6}$ & $5.2_{\pm 0.7}$ & $6.2_{\pm 0.5}$ & $13.6_{\pm 1.5}$ & $12.3_{\pm 1.1}$ \\
    \hdashline
    \multirow{2}{*}{\name} & Performance & \textbf{35.2}$_{\pm \mathbf{0.5}}$/\textbf{32.8}$_{\pm \mathbf{0.4}}$ & \textbf{91.3}$_{\pm \mathbf{0.4}}$ & \textbf{71.0}$_{\pm \mathbf{0.5}}$ & \textbf{63.8}$_{\pm \mathbf{0.5}}$ & \textbf{21.4}$_{\pm \mathbf{0.6}}$ & \textbf{8.5}$_{\pm \mathbf{0.4}}$ \\
    & Tokens (K) & \textbf{7.5}$_{\pm \mathbf{0.7}}$/\textbf{9.1}$_{\pm \mathbf{0.6}}$ & \textbf{3.8}$_{\pm \mathbf{0.4}}$ & \textbf{2.9}$_{\pm \mathbf{0.3}}$ & \textbf{3.1}$_{\pm \mathbf{0.4}}$ & \textbf{11.3}$_{\pm \mathbf{0.5}}$ & \textbf{9.7}$_{\pm \mathbf{1.0}}$ \\
    \bottomrule
    \end{tabular}
    } 
    \vspace{-0.8em}
\end{table}

\subsection{Analysis of Orchestration Abstraction} \label{app:analysis_orchestration} 

To isolate the contribution of the milestone abstraction itself from the learning machinery built on top of it, we remove milestones and instead treat the entire query $q$ as a single milestone. Under this variant, the orchestrator generates a single plan for the complete task, tree branching occurs only at plan decisions, and all other training and execution settings remain identical to \name. This setup directly instantiates the ``plan a complete workflow upfront'' regime discussed in \Cref{sec:introduction}, allowing us to evaluate the effect of the milestone abstraction in isolation.

As shown in \Cref{tab:ablation_orchestration}, removing milestones produces the largest overall degradation among the ablations we consider. Task performance decreases by $6.0$, $5.8$, and $3.9$ points on mathematical reasoning, embodied decision-making, and deep research, respectively. These are the largest drops among all variants in \Cref{tab:ablation} for mathematical reasoning and embodied decision-making, and the second largest for deep research. At the same time, token cost increases by $37\%$, $85\%$, and $27\%$, reducing intelligence efficiency to $0.65$, $0.49$, and $0.57\times$ that of \name. The resulting efficiency is lower than that of any other ablation on embodied decision-making and deep research. Notably, removing the milestone abstraction is more detrimental than replacing tree-structured rollouts or removing any individual reward, indicating that the improvements in \Cref{sec:main_results} cannot be attributed solely to applying the proposed reinforcement learning procedure at an arbitrary decision granularity.

The domain-specific effects further illustrate the role of milestones. Deep research exhibits the largest \emph{relative} performance degradation, at $27\%$, consistent with the motivation that committing to a complete workflow upfront makes it harder to revise the execution strategy when intermediate results reveal new information requirements. Embodied decision-making exhibits the largest increase in token cost, consistent with the benefit of periodically re-scoping execution at milestone boundaries rather than continuing a long-horizon plan after the task state has changed. Finally, removing milestones eliminates milestone-level branching and leaves only plan-level alternatives, thereby reducing the set of decision points at which matched downstream outcomes can be compared. Credit must consequently be assigned over longer spans of execution, directly exacerbating the decision-level credit-assignment challenge that motivates the milestone abstraction.

\begin{table}[ht]
\centering
    \caption{Ablation results of orchestration abstraction, where we remove the milestone structure. We report task performance ($\uparrow$) and token cost ($\downarrow$), the weighted executor output tokens per instance in thousands, their ratio is the intelligence efficiency of \Cref{fig:efficiency}.}
    \label{tab:ablation_orchestration}
    \resizebox{0.85\linewidth}{!}{
    \begin{tabular}{lcccccc}
    \toprule
    \multirow{2}{*}{\textbf{Method}} & \multicolumn{2}{c}{\cellcolor{cyan!20} \textbf{Mathematical}} & \multicolumn{2}{c}{\cellcolor{Orchid!20} \textbf{Embodied}} & \multicolumn{2}{c}{\cellcolor{LimeGreen!20} \textbf{Deep Research}} \\
    & \textbf{Perf.} & \textbf{Tokens (K)} & \textbf{Perf.} & \textbf{Tokens (K)} & \textbf{Perf.} & \textbf{Tokens (K)} \\
    \midrule
    \name & \textbf{52.3}$_{\pm \mathbf{0.4}}$ & \textbf{6.2}$_{\pm \mathbf{0.5}}$ & \textbf{66.5}$_{\pm \mathbf{0.5}}$ & \textbf{3.3}$_{\pm \mathbf{0.4}}$ & \textbf{14.2}$_{\pm \mathbf{0.4}}$ & \textbf{10.4}$_{\pm \mathbf{0.4}}$ \\
    \midrule
    \textit{w/o} Milestones & $46.3_{\pm 0.4}$ & $8.5_{\pm 0.7}$ & $60.7_{\pm 0.5}$ & $6.1_{\pm 0.4}$ & $10.3_{\pm 0.3}$ & $13.2_{\pm 0.7}$ \\
    \bottomrule
    \end{tabular}
    } 
\end{table} 

\subsection{Ablation of Completeness Reward} \label{app:ablation_completeness} 

The completeness reward is the only component that directly targets the structural validity of an orchestration decision rather than its downstream outcome. We therefore evaluate its effect through plan validity rather than task performance. 
Following the definition in \Cref{sec:reward_design}, we consider four criteria: whether a plan is successfully parsed with all required fields, whether its dependencies form a directed acyclic graph (DAG), whether every step is assigned to a valid executor in the pool, and whether the resulting plan can be executed to produce a milestone-level outcome.

\begin{table}[ht]
    \centering
    \caption{Ablation of the completeness reward. We report the percentage of orchestration decisions passing each validity check, measured over all plans generated during evaluation on all benchmarks. Format and assignment validity are syntactic checks, DAG validity is structural, and \textbf{Executable Plan} requires all three together with successful production of a milestone-level outcome.}
    \label{tab:ablation_completeness}
    \resizebox{0.85\linewidth}{!}{
    \begin{tabular}{lcccc}
    \toprule
    \textbf{Method} & \textbf{Valid Format ↑} & \textbf{Valid DAG ↑} & \textbf{Valid Assignment ↑} & \textbf{Executable Plan ↑} \\
    \midrule
    \name & $100.0$ & $96.3$ & $100.0$ & $95.4$ \\
    \textit{w/o} Completeness Reward & $74.3$ & $62.1$ & $85.7$ & $60.4$ \\
    \bottomrule
    \end{tabular}
    }
\end{table}

As shown in \Cref{tab:ablation_completeness}, \name achieves perfect format and executor-assignment validity, while its remaining failures arise primarily from malformed dependency structures, with $3.7\%$ of plans violating DAG validity. 
Removing the completeness reward degrades all four criteria, but to markedly different extents. 
Executor-assignment validity decreases by $14.3$ percentage points, whereas DAG validity and overall executability decrease by $34.2$ and $35.0$ points, respectively. 
This difference reflects the nature of the underlying constraints: selecting a valid executor is largely a local decision, facilitated by explicitly enumerating the admissible executor identifiers in the prompt, whereas constructing an acyclic dependency graph imposes a global structural constraint over all steps in the plan. 
The ablation results therefore indicate that prompting alone is insufficient to reliably enforce such global structural consistency.

The impact of completeness extends beyond output validity. 
A non-executable plan produces no milestone-level outcome and consequently provides no downstream signal from either the performance or cost reward. 
With only $60.4\%$ of plans executable after removing the completeness reward, roughly two in five orchestration decisions fail to yield usable downstream feedback. 
This also reduces the number of valid sibling branches available for the within-sibling comparisons underlying the other two rewards. 
The completeness reward thus serves as a structural safeguard for decision-level learning, helping preserve the valid rollout branches required by the performance and cost objectives. 
Notably, the completeness reward remains effective with a relatively small weight ($\lambda_{\mathrm{comp}}=0.1$), while removing it entirely leads to substantially more invalid orchestration decisions.

\subsection{Case Studies}\label{app:case_study}

\paragraph{Performance-Oriented Orchestration.}
The paired AIME25 examples in Case~\ref{cs:aime25_case} and Case~\ref{cs:baseline_aime25_case} illustrate the role of milestone decomposition and dependency-aware planning. 
\name decomposes the problem into three verifiable milestones---deriving the divisibility constraints, counting valid assignments, and computing the requested difference---and specifies an executor and dependency list for every step. 
The resulting plan forms an executable DAG. 
By contrast, the baseline trace contains a missing dependency field and a circular dependency, making the plan unreliable to execute. 
In this pair of traces, the structured milestone-plan-act interface avoids the dependency and format errors present in the baseline plan. We draw the comparison from these two examples alone and do not quantify how often such errors occur.

\paragraph{Efficiency-Oriented Orchestration.}
The paired GAIA examples in Case~\ref{cs:gaia_case} and Case~\ref{cs:baseline_gaia_case} illustrate capability--cost-aware executor assignment. 
\name delegates straightforward retrieval and extraction to the lower-cost \texttt{model\_a}, while using the stronger \texttt{model\_b} for the consistency check needed to avoid an entity mismatch. 
The baseline trace reaches the same final answer but repeatedly invokes \texttt{model\_b} for simple or redundant checks and adds an unnecessary summarization step. 
This contrast is consistent with the intended behavior of the learned policy: allocate stronger executors only when their additional capability is useful, and otherwise favor lower-cost execution.

\casepageseparator
\begin{case}{\name Case on AIME25}{aime25_case}
\begin{casesummary}
\textbf{Task.} There are $8! = 40320$ eight-digit positive integers that use each of the digits $1, 2, 3, 4, 5, 6, 7, 8$ exactly once. Let $N$ be the number of these integers that are divisible by $22$. Find the difference between $N$ and $2025$.

\smallskip
\textbf{Trace.} 3 milestones; 10 planned steps. \textbf{Assignments.} \texttt{model\_a}: 7; \texttt{model\_b}: 3.
\end{casesummary}
\casemilestone{Milestone 1. Derive the divisibility constraints for $22$.}
\begingroup
\small
\renewcommand{\arraystretch}{1.10}
\setlength{\tabcolsep}{4pt}
\begin{tabularx}{\linewidth}{@{}>{\centering\arraybackslash}p{0.055\linewidth} >{\raggedright\arraybackslash}X >{\centering\arraybackslash}p{0.13\linewidth} >{\centering\arraybackslash}p{0.13\linewidth}@{}}
\toprule
\textbf{Step} & \textbf{Sub-task} & \textbf{Executor} & \textbf{Depends on} \\
\midrule
1 & Factor $22$ into $2 \times 11$ and identify the required divisibility conditions. & \caseexecutora & $\emptyset$ \\
2 & Apply the divisibility rule for $11$ to the digit positions. & \caseexecutorb & $[1]$ \\
3 & Verify that the sums of digits in odd and even positions must be equal, and hence both must be $18$. & \caseexecutora & $[2]$ \\
4 & Verify that the last digit must be even to satisfy divisibility by $2$. & \caseexecutora & $[1]$ \\
\bottomrule
\end{tabularx}
\endgroup
\begin{caseoutcome}
\textbf{Outcome.} A valid number must satisfy two constraints: the odd-position digits and even-position digits each sum to $18$, and the last digit is even.
\end{caseoutcome}

\casemilestone{Milestone 2. Count the valid digit assignments satisfying the derived constraints.}
\begingroup
\small
\renewcommand{\arraystretch}{1.10}
\setlength{\tabcolsep}{4pt}
\begin{tabularx}{\linewidth}{@{}>{\centering\arraybackslash}p{0.055\linewidth} >{\raggedright\arraybackslash}X >{\centering\arraybackslash}p{0.13\linewidth} >{\centering\arraybackslash}p{0.13\linewidth}@{}}
\toprule
\textbf{Step} & \textbf{Sub-task} & \textbf{Executor} & \textbf{Depends on} \\
\midrule
1 & Enumerate all $4$-digit subsets of $\{1, 2, 3, 4, 5, 6, 7, 8\}$ whose sum is $18$. & \caseexecutorb & $\emptyset$ \\
2 & Count the number of arrangements of each valid subset in the odd positions. & \caseexecutora & $[1]$ \\
3 & Count the number of valid arrangements of the complementary digits in the even positions under the constraint that the last digit is even. & \caseexecutora & $[1]$ \\
4 & Aggregate the counts from odd-position and even-position arrangements to obtain $N$. & \caseexecutorb & $[2, 3]$ \\
\bottomrule
\end{tabularx}
\endgroup
\begin{caseoutcome}
\textbf{Outcome.} There are $8$ valid choices for the odd-position digit set. For each choice, there are $4!$ arrangements of the odd positions and $2 \times 3!$ valid arrangements of the even positions. Thus,
\[
N = 8 \times 4! \times 2 \times 3! = 2304.
\]
\end{caseoutcome}

\casemilestone{Milestone 3. Compute the final requested difference.}
\begingroup
\small
\renewcommand{\arraystretch}{1.10}
\setlength{\tabcolsep}{4pt}
\begin{tabularx}{\linewidth}{@{}>{\centering\arraybackslash}p{0.055\linewidth} >{\raggedright\arraybackslash}X >{\centering\arraybackslash}p{0.13\linewidth} >{\centering\arraybackslash}p{0.13\linewidth}@{}}
\toprule
\textbf{Step} & \textbf{Sub-task} & \textbf{Executor} & \textbf{Depends on} \\
\midrule
1 & Compute $2304 - 2025$. & \caseexecutora & $\emptyset$ \\
2 & Return the final answer. & \caseexecutora & $[1]$ \\
\bottomrule
\end{tabularx}
\endgroup
\begin{caseoutcome}
\textbf{Final answer.} $\boxed{279}$
\end{caseoutcome}
\begin{caseanalysis}
The trace separates constraint derivation, counting, and final arithmetic into three milestones. Executor \texttt{model\_b} handles the divisibility argument, subset enumeration, and count aggregation, while \texttt{model\_a} performs local checks and arithmetic. Explicit dependencies expose independent checks and keep the counting steps ordered. The recorded final answer is $279$.
\end{caseanalysis}
\end{case}

\casepageseparator
\begin{case}{Baseline Case on AIME25}{baseline_aime25_case}
\begin{casesummary}
\textbf{Task.} There are $8! = 40320$ eight-digit positive integers that use each of the digits $1, 2, 3, 4, 5, 6, 7, 8$ exactly once. Let $N$ be the number of these integers that are divisible by $22$. Find the difference between $N$ and $2025$.

\smallskip
\textbf{Trace.} 1 milestone; 5 planned steps. \textbf{Assignments.} \texttt{model\_a}: 3; \texttt{model\_b}: 2.
\end{casesummary}
\casemilestone{Milestone 1. Solve the divisibility and counting problem.}
\begingroup
\small
\renewcommand{\arraystretch}{1.10}
\setlength{\tabcolsep}{4pt}
\begin{tabularx}{\linewidth}{@{}>{\centering\arraybackslash}p{0.055\linewidth} >{\raggedright\arraybackslash}X >{\centering\arraybackslash}p{0.13\linewidth} >{\centering\arraybackslash}p{0.13\linewidth}@{}}
\toprule
\textbf{Step} & \textbf{Sub-task} & \textbf{Executor} & \textbf{Depends on} \\
\midrule
1 & Directly solve the full problem by deriving the divisibility rules, counting valid permutations, and computing the final answer. & \caseexecutorb & $\emptyset$ \\
2 & Re-derive the divisibility rule for $22$ and verify the counting result. & \caseexecutorb & $\emptyset$ \\
3 & Count the number of valid digit assignments satisfying the divisibility constraints. & \caseexecutora & $[4]$ \\
4 & Summarize the intermediate results and return the final answer. & \caseexecutora & $[3]$ \\
5 & Check whether the final answer is reasonable. & \caseexecutora & \textit{Missing} \\
\bottomrule
\end{tabularx}
\endgroup

\begin{caseanalysis}
The plan contains a dependency cycle between Steps 3 and 4, while Step 5 omits the required dependency field. These structural defects prevent the plan from defining a valid executable DAG. Steps 1 and 2 both use \texttt{model\_b} for broad, overlapping reasoning, rather than assigning smaller verifiable sub-tasks. No final answer is included in the supplied trace.
\end{caseanalysis}
\end{case}

\casepageseparator
\begin{case}{\name Case on GAIA}{gaia_case}
\begin{casesummary}
\textbf{Task.} The Metropolitan Museum of Art has a portrait in its collection with an accession number of $29.100.5$. Of the consecrators and co-consecrators of this portrait's subject as a bishop, what is the name of the one who never became pope?

\smallskip
\textbf{Trace.} 3 milestones; 9 planned steps. \textbf{Assignments.} \texttt{model\_a}: 7; \texttt{model\_b}: 2.
\end{casesummary}
\casemilestone{Milestone 1. Identify the portrait subject associated with accession number $29.100.5$.}
\begingroup
\small
\renewcommand{\arraystretch}{1.10}
\setlength{\tabcolsep}{4pt}
\begin{tabularx}{\linewidth}{@{}>{\centering\arraybackslash}p{0.055\linewidth} >{\raggedright\arraybackslash}X >{\centering\arraybackslash}p{0.13\linewidth} >{\centering\arraybackslash}p{0.13\linewidth}@{}}
\toprule
\textbf{Step} & \textbf{Sub-task} & \textbf{Executor} & \textbf{Depends on} \\
\midrule
1 & Search the Metropolitan Museum of Art collection using accession number \texttt{29.100.5}. & \caseexecutora & $\emptyset$ \\
2 & Extract the artwork title and portrait subject from the collection page. & \caseexecutora & $[1]$ \\
\bottomrule
\end{tabularx}
\endgroup
\begin{caseoutcome}
\textbf{Outcome.} The portrait subject is Cardinal Fernando Niño de Guevara.
\end{caseoutcome}

\casemilestone{Milestone 2. Retrieve the consecrators and co-consecrators of Fernando Niño de Guevara.}
\begingroup
\small
\renewcommand{\arraystretch}{1.10}
\setlength{\tabcolsep}{4pt}
\begin{tabularx}{\linewidth}{@{}>{\centering\arraybackslash}p{0.055\linewidth} >{\raggedright\arraybackslash}X >{\centering\arraybackslash}p{0.13\linewidth} >{\centering\arraybackslash}p{0.13\linewidth}@{}}
\toprule
\textbf{Step} & \textbf{Sub-task} & \textbf{Executor} & \textbf{Depends on} \\
\midrule
1 & Search for the episcopal consecration record of Fernando Niño de Guevara. & \caseexecutora & $\emptyset$ \\
2 & Extract the principal consecrator and co-consecrators from the retrieved record. & \caseexecutora & $[1]$ \\
3 & Verify that the extracted consecration record corresponds to the portrait subject to avoid entity mismatch. & \caseexecutorb & $[2]$ \\
\bottomrule
\end{tabularx}
\endgroup
\begin{caseoutcome}
\textbf{Outcome.} Fernando Niño de Guevara was consecrated by Pope Clement VIII, with Camillo Borghese and Alfonso Visconti serving as co-consecrators.
\end{caseoutcome}

\casemilestone{Milestone 3. Determine which consecrator or co-consecrator never became pope.}
\begingroup
\small
\renewcommand{\arraystretch}{1.10}
\setlength{\tabcolsep}{4pt}
\begin{tabularx}{\linewidth}{@{}>{\centering\arraybackslash}p{0.055\linewidth} >{\raggedright\arraybackslash}X >{\centering\arraybackslash}p{0.13\linewidth} >{\centering\arraybackslash}p{0.13\linewidth}@{}}
\toprule
\textbf{Step} & \textbf{Sub-task} & \textbf{Executor} & \textbf{Depends on} \\
\midrule
1 & Verify that Clement VIII was pope. & \caseexecutora & $\emptyset$ \\
2 & Verify that Camillo Borghese later became Pope Paul V. & \caseexecutora & $\emptyset$ \\
3 & Identify the remaining consecrator or co-consecrator and verify whether he ever became pope. & \caseexecutorb & $[1, 2]$ \\
4 & Return the final answer. & \caseexecutora & $[3]$ \\
\bottomrule
\end{tabularx}
\endgroup
\begin{caseoutcome}
\textbf{Final answer.} \textbf{Alfonso Visconti}
\end{caseoutcome}
\begin{caseanalysis}
The trace uses \texttt{model\_a} for collection lookup, extraction, and straightforward status checks. It invokes \texttt{model\_b} twice: to check the identity of the consecration record and to verify the remaining candidate. The first two checks in Milestone 3 have no dependencies and can run independently. The recorded answer matches the baseline trace, with fewer steps and fewer assignments to \texttt{model\_b}.
\end{caseanalysis}
\end{case}

\casepageseparator
\begin{case}{Baseline Case on GAIA}{baseline_gaia_case}
\begin{casesummary}
\textbf{Task.} The Metropolitan Museum of Art has a portrait in its collection with an accession number of $29.100.5$. Of the consecrators and co-consecrators of this portrait's subject as a bishop, what is the name of the one who never became pope?

\smallskip
\textbf{Trace.} 3 milestones; 11 planned steps. \textbf{Assignments.} \texttt{model\_a}: 4; \texttt{model\_b}: 7.
\end{casesummary}
\casemilestone{Milestone 1. Identify the portrait subject and collect background information.}
\begingroup
\small
\renewcommand{\arraystretch}{1.10}
\setlength{\tabcolsep}{4pt}
\begin{tabularx}{\linewidth}{@{}>{\centering\arraybackslash}p{0.055\linewidth} >{\raggedright\arraybackslash}X >{\centering\arraybackslash}p{0.13\linewidth} >{\centering\arraybackslash}p{0.13\linewidth}@{}}
\toprule
\textbf{Step} & \textbf{Sub-task} & \textbf{Executor} & \textbf{Depends on} \\
\midrule
1 & Search the Metropolitan Museum of Art collection using accession number \texttt{29.100.5} and identify the artwork title, artist, date, and portrait subject. & \caseexecutorb & $\emptyset$ \\
2 & Search additional sources to reconfirm the same accession number, artwork title, and portrait subject. & \caseexecutorb & $[1]$ \\
3 & Extract the final portrait subject from the verified collection information. & \caseexecutora & $[2]$ \\
\bottomrule
\end{tabularx}
\endgroup
\begin{caseoutcome}
\textbf{Outcome.} The portrait subject is Cardinal Fernando Niño de Guevara.
\end{caseoutcome}

\casemilestone{Milestone 2. Retrieve the episcopal consecration information for Fernando Niño de Guevara.}
\begingroup
\small
\renewcommand{\arraystretch}{1.10}
\setlength{\tabcolsep}{4pt}
\begin{tabularx}{\linewidth}{@{}>{\centering\arraybackslash}p{0.055\linewidth} >{\raggedright\arraybackslash}X >{\centering\arraybackslash}p{0.13\linewidth} >{\centering\arraybackslash}p{0.13\linewidth}@{}}
\toprule
\textbf{Step} & \textbf{Sub-task} & \textbf{Executor} & \textbf{Depends on} \\
\midrule
1 & Search for the episcopal consecration record of Fernando Niño de Guevara and extract the principal consecrator and co-consecrators. & \caseexecutorb & $\emptyset$ \\
2 & Independently search another source for the same consecration record and compare the extracted names. & \caseexecutorb & $[1]$ \\
3 & Merge the consecration information into a single list of names. & \caseexecutora & $[1, 2]$ \\
\bottomrule
\end{tabularx}
\endgroup
\begin{caseoutcome}
\textbf{Outcome.} Fernando Niño de Guevara was consecrated by Pope Clement VIII, with Camillo Borghese and Alfonso Visconti serving as co-consecrators.
\end{caseoutcome}

\casemilestone{Milestone 3. Determine which consecrator or co-consecrator never became pope.}
\begingroup
\small
\renewcommand{\arraystretch}{1.10}
\setlength{\tabcolsep}{4pt}
\begin{tabularx}{\linewidth}{@{}>{\centering\arraybackslash}p{0.055\linewidth} >{\raggedright\arraybackslash}X >{\centering\arraybackslash}p{0.13\linewidth} >{\centering\arraybackslash}p{0.13\linewidth}@{}}
\toprule
\textbf{Step} & \textbf{Sub-task} & \textbf{Executor} & \textbf{Depends on} \\
\midrule
1 & Verify the papal status of Pope Clement VIII. & \caseexecutorb & $\emptyset$ \\
2 & Verify whether Camillo Borghese later became pope. & \caseexecutorb & $\emptyset$ \\
3 & Verify whether Alfonso Visconti ever became pope. & \caseexecutorb & $\emptyset$ \\
4 & Identify the consecrator or co-consecrator who never became pope from the verified status results. & \caseexecutora & $[1, 2, 3]$ \\
5 & Summarize the evidence and return the final answer. & \caseexecutora & $[4]$ \\
\bottomrule
\end{tabularx}
\endgroup
\begin{caseoutcome}
\textbf{Outcome.} Clement VIII was pope, Camillo Borghese later became Pope Paul V, and Alfonso Visconti did not become pope.

\textbf{Final answer.} \textbf{Alfonso Visconti}
\end{caseoutcome}
\begin{caseanalysis}
The trace assigns seven of its eleven steps to \texttt{model\_b}, including collection lookup, repeated retrieval, and all three papal-status checks. The supplied annotations identify redundant confirmation in Milestone 1, duplicate retrieval in Milestone 2, and an unnecessary final summarization step. The extra dependency in Milestone 1 also delays extraction. The trace reaches the same recorded answer as the corresponding \name case.
\end{caseanalysis}
\end{case}

\section{Limitations and Future Work}
\label{app:limitations}

Our prompts, executor profiles, and evaluation benchmarks are predominantly English. The current results therefore do not establish whether the learned orchestration policy transfers to multilingual tasks, code-switched inputs, or language-specific tools and executors. Multilingual settings may require the orchestrator to account explicitly for language proficiency and cross-lingual information integration.

The current evaluation further considers a limited number of executor configurations and assumes that the supplied capability and cost profiles are accurate and stable. In practice, executor behavior may vary across domains, prompts, and deployment conditions, while larger agent pools may substantially increase the orchestration search space. Evaluating robustness to noisy or evolving profiles and scaling to larger, more heterogeneous executor pools remain important directions for future work.

Finally, tree-structured rollouts increase training-time computation. The number of candidate trajectories grows with the number of initial trajectories ($B$), branches per expansion ($N$), and expansion rounds ($E$). Selective branching, early pruning, and trajectory reuse may reduce this training overhead.

\end{document}